\documentclass[10pt,twocolumn,letterpaper]{article}

\usepackage{iccv}

\makeatletter
\@namedef{ver@everyshi.sty}{}
\makeatother
\usepackage{tikz}

\usepackage{times}
\usepackage{epsfig}
\usepackage{graphicx}
\usepackage{amsmath}
\usepackage{amssymb}

\usepackage[pagebackref=true,breaklinks=true,letterpaper=true,colorlinks,bookmarks=false]{hyperref}

\def\iccvPaperID{2913} 

 \iccvfinalcopy

\ificcvfinal\fi

\usepackage{array}
\newcolumntype{H}{>{\setbox0=\hbox\bgroup}c<{\egroup}@{}}
\usepackage[export]{adjustbox}
\usepackage{multirow}
\usepackage{todonotes} 

\usepackage{subfigure}
\usepackage{cuted}
\usepackage{capt-of}

\usepackage{booktabs}  

\usepackage{appendix}

\begin{document}

\title{Boosting 3-DoF Ground-to-Satellite Camera Localization Accuracy via Geometry-Guided Cross-View Transformer}

\author{Yujiao Shi\textsuperscript{\rm 1}, Fei Wu\textsuperscript{\rm 1},  Akhil Perincherry\textsuperscript{\rm 2},  Ankit Vora\textsuperscript{\rm 2} and Hongdong Li\textsuperscript{\rm 1}\\
\textsuperscript{\rm 1}The Australian National University~~
\textsuperscript{\rm 2}Ford Motor Company\\
{\tt\small yujiao.shi@anu.edu.au}
}

\maketitle
\ificcvfinal\thispagestyle{empty}\fi

\begin{abstract}


Image retrieval-based cross-view localization methods often lead to very coarse camera pose estimation, due to the limited sampling density of the database satellite images. In this paper, we propose a method to increase the accuracy of a ground camera's location and orientation by estimating the relative rotation and translation between the ground-level image and its matched/retrieved satellite image.
Our approach designs a geometry-guided cross-view transformer that combines the benefits of conventional geometry and learnable cross-view transformers to map the ground-view observations to an overhead view. 
Given the synthesized overhead view and observed satellite feature maps, we construct a neural pose optimizer with strong global information embedding ability to estimate the relative rotation between them. After aligning their rotations, we develop an uncertainty-guided spatial correlation to generate a probability map of the vehicle locations, from which the relative translation can be determined.
Experimental results demonstrate that our method significantly outperforms the state-of-the-art. Notably, the likelihood of restricting the vehicle lateral pose to be within 1m of its Ground Truth (GT) value on the cross-view KITTI dataset has been improved from $35.54\%$ to $76.44\%$, and the likelihood of restricting the vehicle orientation to be within $1^{\circ}$ of its GT value has been improved from $19.64\%$ to $99.10\%$.

\end{abstract}

\begin{figure}[ht!]
    \centering
    \begin{minipage}{0.44\linewidth}
    \centering
            \adjincludegraphics[width=0.9\linewidth,trim={{0.22\width} {0.22\width} {0.22\width} {0.22\width}},clip]{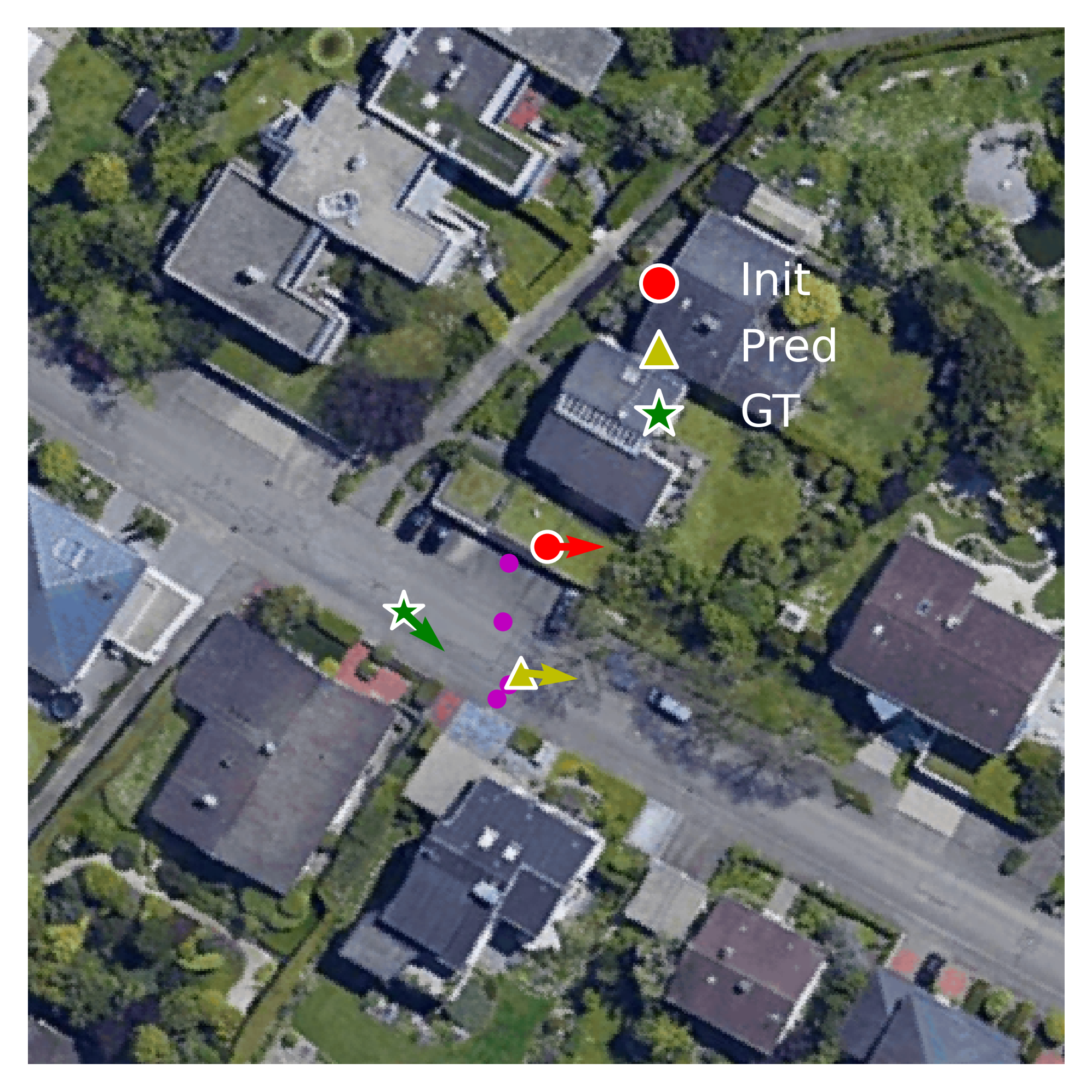}\\
            \adjincludegraphics[width=0.9\linewidth,trim={{0.22\width} {0.22\width} {0.22\width} {0.22\width}},clip]{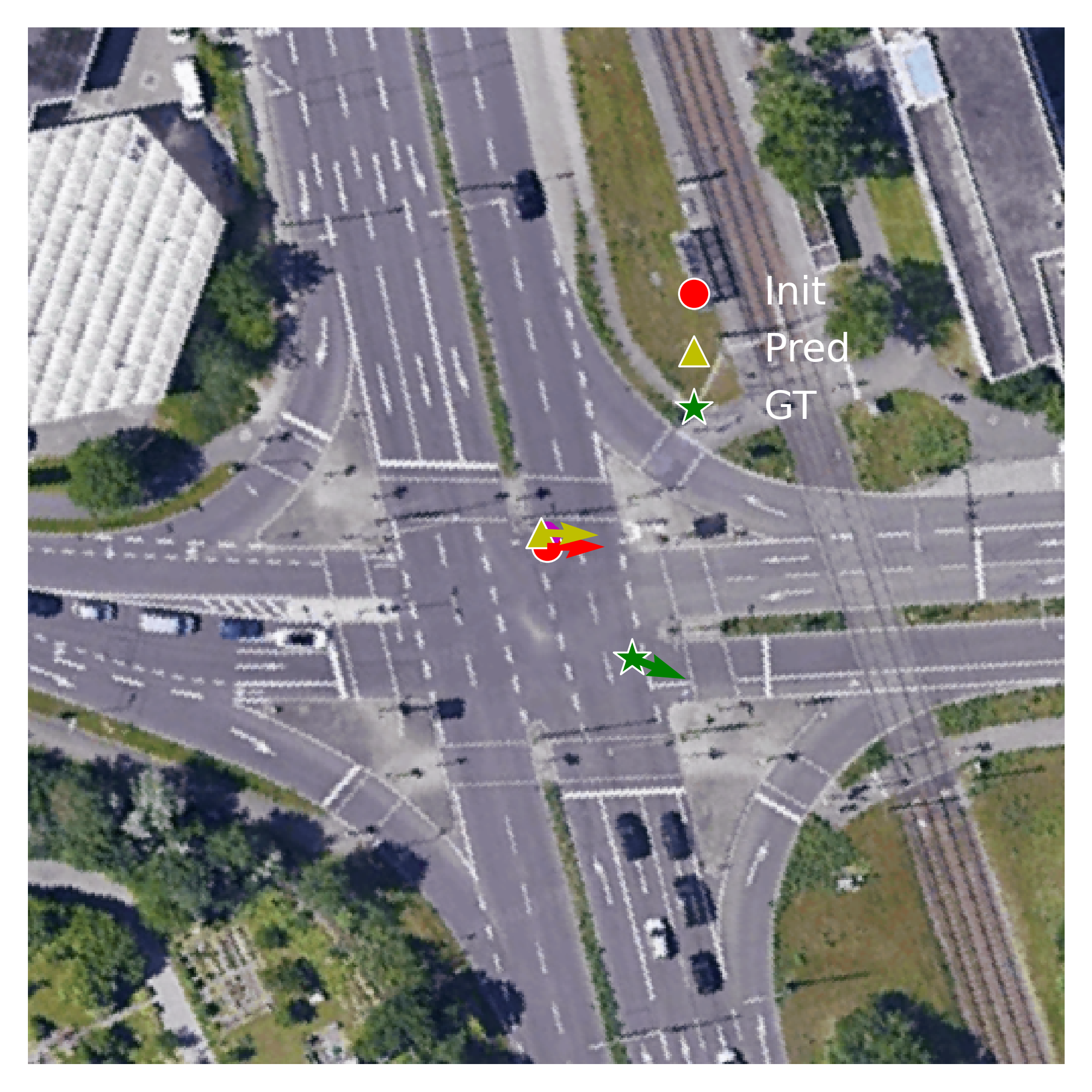}\\
            \text{\footnotesize (a) Joint 3-DoF pose {optimization}~\cite{shi2022beyond}}
    \end{minipage}
    \hspace{1em}
    \begin{minipage}{0.44\linewidth}
\centering
     \adjincludegraphics[width=0.9\linewidth,trim={{0.23\width} {0.23\width} {0.23\width} {0.23\width}},clip]{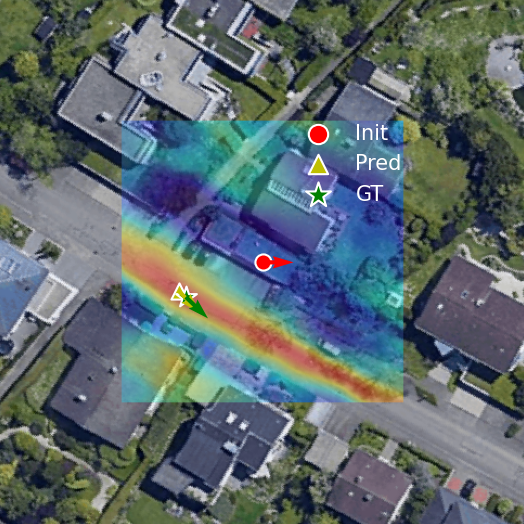}\\
    \adjincludegraphics[width=0.9\linewidth,trim={{0.23\width} {0.23\width} {0.23\width} {0.23\width}},clip]{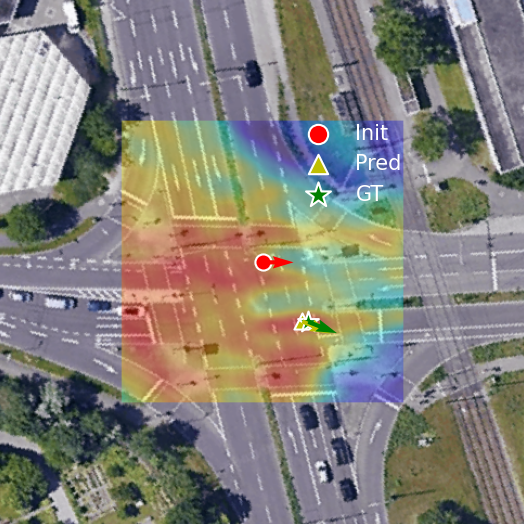}\\
    \text{\footnotesize (b) Ours} \\
    
\end{minipage}
\caption{
Conventional image-retrieval methods for ground-to-satellite localization provide a rough location and orientation estimation for the ground camera, denoted by the red dot and arrow in the images. 
This paper aims to refine this pose under the same ground-to-satellite image-matching context. 
Compared to joint rotation and translation \textit{optimization}~\cite{shi2022beyond}, which is susceptible to local minima (a), this paper introduces a new method that allows \textit{dense search} at all possible locations. 
Our method produces a probability map (b) over the continuous search space of vehicle locations, thus achieving high localization accuracy. 
}
\label{fig:teaser} 
\end{figure}

\section{Introduction}

Autonomous robots, such as unmanned aerial vehicles (UAVs) and unmanned ground vehicles (UGVs), are becoming more and more popular in various fields of applications.
One of the most demanding capabilities of autonomous vehicles is navigating and executing tasks autonomously in complex environments, especially in environments with poor GPS signals. 
This motivates recent research on vision-based localization. 

Ground-to-satellite image-based localization is a vision-based localization task that aims to estimate the location and orientation of a ground camera by matching a ground-level image against a large satellite map. 
The task was originally proposed for city-scale localization and tackled by image retrieval techniques. 
However, image retrieval techniques can only provide a rough pose approximation of the ground camera, and the sample density of the database image always dictates the estimated pose accuracy.

Recent works have explored increasing the localization accuracy by estimating a relative translation and rotation between the ground and its matching satellite images. 
Existing works have tried to regress the relative translation by MLPs~\cite{zhu2021vigor}, or further split the retrieved satellite image to a $N\times N$ grid and match the ground image against the grid to improve the localization accuracy~\cite{xia2022visual}. 
However, both of them cannot estimate the orientation of ground cameras. 
To estimate a 3-DoF camera pose (location and orientation), a deep cross-view optimization scheme has been proposed~\cite{shi2022beyond}. 
Nonetheless, optimization-based methods are highly susceptible to local minima, as shown in Fig.~\ref{fig:teaser} (a). 

To avoid this issue, this paper presents a new framework that allows searching for vehicle locations densely over the entire solution space. 
It estimates the orientation and location of the ground camera sequentially and is designed based on an observation that neural networks 
behave differently to rotations and translations on input signals. 

Specifically, neural network outputs tend to magnify the rotation difference on input signals, making it a desired choice for rotation estimation. 
Thus, we propose a neural network-based pose optimizer. It works with an overhead view feature synthesis module, which synthesizes an overhead view feature map from the query ground view image, to estimate the relative rotation between the ground and satellite image. 
However, due to feature aggregation layers (\eg, pooling), a small translation difference in input signals might be absorbed in high-level deep features. This makes the estimated translation by an optimizer constructed by neural networks inaccurate. 

On the other hand, when the orientation of the synthesized overhead view feature map from the ground view image has been aligned with the satellite image, the vehicle location can be obtained by performing a spatial correlation between them. 
The spatial correlation generates a probability map of vehicle locations over the entire search space, as shown in Fig.~\ref{fig:teaser} (b). This allows a dense and exhaustive search for vehicle locations.

Our overhead view feature synthesis module is designed as a geometry-guided cross-view transformer. It explicitly embeds the deterministic geometric correspondences to the learnable cross-view transformers. 
Compared to the ground plane homography projection in \cite{shi2022beyond}, our method handles the height ambiguity of scene objects and the ground camera's slight tilt and roll angle change.

Our contributions are summarized as follows: (1) a rotation and translation decoupled cross-view camera localization framework, which achieves state-of-the-art performance on four widely used benchmarks;
(2) a neural pose optimizer for rotation estimation, which produces highly-accurate rotation estimation results;
(3) a dense search mechanism for translation estimation, which computes a possibility map of vehicle locations over the entire search space;  
(4) a geometry-guided cross-view transformer for ground-to-overhead view feature synthesis, which combines the wisdom of deterministic geometric and data-driven learnable correspondences. 


\begin{figure*}[ht!]
    \centering
    \includegraphics[width=\linewidth]{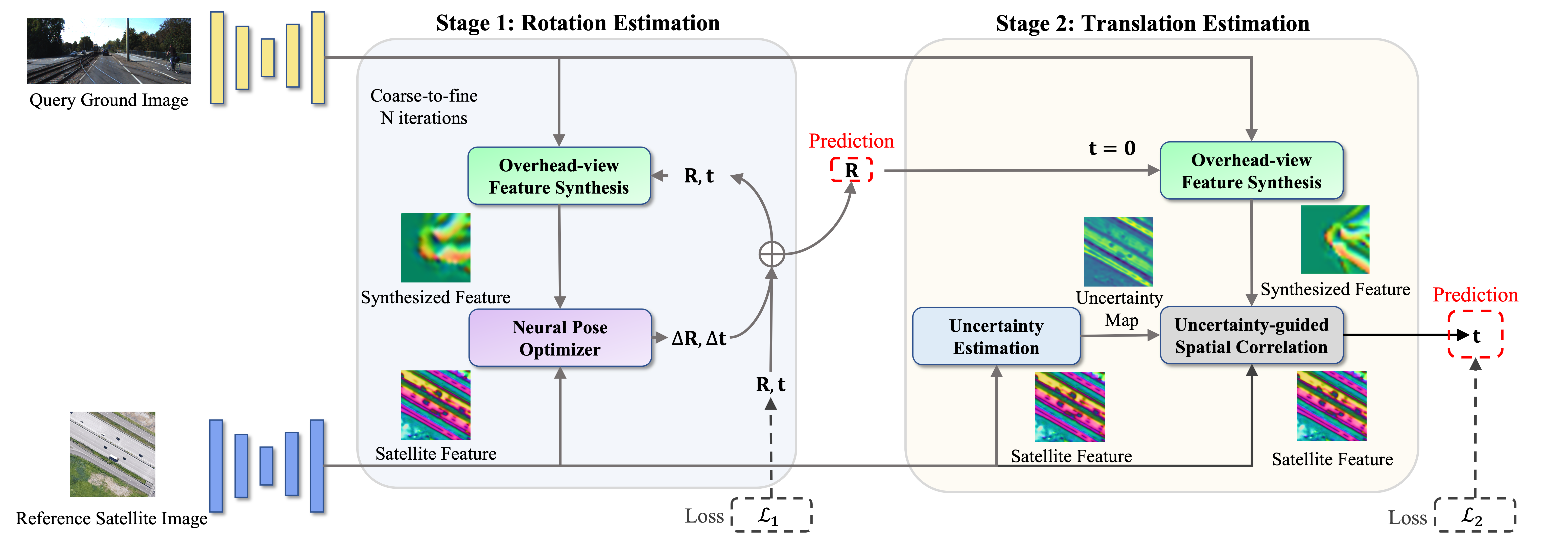}
    \caption{An overview of the proposed method. 
    (Stage 1:) We design an overhead-view feature synthesis module to map ground-view image features to the overhead view, according to a relative rotation $\mathbf{R}$ and translation $\mathbf{t}$. 
    Taking input as the feature differences between the synthesized and observed overhead view features, our proposed neural pose optimizer updates the relative transformation from coarse to finer feature levels. 
    (Stage 2:) Then, we re-synthesize an overhead view feature map according to the final estimated $\mathbf{R}$ and zero translation, and use it as a sliding window to compute its spatial correlation with the observed satellite feature map. 
    We also estimate an uncertainty map from the satellite semantic features. The uncertainty map is encoded in the spatial correlation process to exclude impossible camera locations, \eg, building and tree areas. 
    The pixel coordinate with the maximum correlation result determines the query camera location. Our neural optimizer's final estimated $\mathbf{R}$ (from stage 1) is regarded as the camera orientation.}
    \label{fig:framework}
\end{figure*}

\section{Related Works}

\textbf{Satellite image-based localization.}
Satellite image-based localization aims to estimate the location and orientation of a ground sensor mounted on a robot or a vehicle by a large satellite image. 
Various works have tried localizing lidar \cite{vora2020aerial, mishra2022infra, veronese2015aerial} or radar \cite{tang2020rsl, tang2021self} points on satellite imagery. 
However, equipping a lidar or radar sensor on a robot is usually expensive. 
In contrast, cameras provide a cheaper option than lidar and radar sensors. 
Thus, using ground-to-satellite image matching for localization has recently attracted tremendous attention.

Ground-to-satellite image-based localization was initially proposed for city-scale localization. 
The task is to retrieve the most similar satellite image from a database to determine the query camera location. 
Conventional works have focused on designing powerful handcrafted features to match the cross-view images~\cite{castaldo2015semantic, lin2013cross, mousavian2016semantic}. 
To handle significant cross-view differences, recent deep metric learning techniques provide a powerful alternative for the cross-view image matching task. 
Researchers have devoted themselves to designing powerful networks~\cite{workman2015location, workman2015wide, vo2016localizing, Cai_2019_ICCV, yang2021cross, Zhu_2022_CVPR}, learning orientation invariant or equivariant descriptors~\cite{Hu_2018_CVPR, Liu_2019_CVPR, sun2019geocapsnet, shi2020looking, zhu2021revisiting}, and bridging the cross-view domain gaps~\cite{zhai2017predicting, Regmi_2019_ICCV, shi2019spatial, shi2020optimal, toker2021coming}. 
However, image retrieval-based methods suffer from poor localization accuracy as they approximate the GPS of retrieved satellite image as the query camera location. 

Recently, researchers have demonstrated that it is possible to accurately localize which pixel on the satellite image corresponds to the query camera location. 
Zhu~\etal~\cite{zhu2021vigor} employ BlackBox MLPs to regress the relative location coordinates. 
Xia~\etal~\cite{xia2022visual} propose a patch-matching method to estimate the uncertainty of the ground vehicle location on a satellite image. 
However, the two works cannot estimate the camera's orientation with respect to the satellite image. 
Considering the large search space of 3-DoF camera pose, Shi and Li~\cite{shi2022beyond} design a deep optimization mechanism to update camera pose iteratively. 

There are also two contemporary works on arXiv~\cite{lentsch2022slicematch, fervers2022uncertainty}. 
Lentsch \etal~\cite{lentsch2022slicematch} propose to generate a number of candidate poses and their corresponding masks on the satellite image. 
The satellite features selected by these masks are matched to the query ground image to determine its pose. 
Instead of sampling discretized poses, our method produces continuous rotation estimates, and our translation is searched uniformly over the entire search space without being affected by the sample randomness. 
Fervers \etal~\cite{fervers2022uncertainty} shares the same insight as us on translation estimation. 
Compared to their method, we encode an uncertainty map in the spatial correlation for translation estimation. The uncertainty map excludes improbable vehicle locations, \eg, areas indicated by buildings or tree canopy.  Furthermore, their rotations are sampled at discretized values, while our method estimates continuous rotations.   

Decoupling rotation and translation has also been explored in other tasks, \eg, absolute pose regression~\cite{kendall2017geometric}, Visual Odometry (VO)~\cite{kim2018low}, and 6-DoF object pose estimation~\cite{chen2021fs}. 
This paper shows that developing different rotation and translation estimation strategies also facilitates the overall performance of ground-to-satellite camera localization. 
Our proposed method tackles the unique challenges of this task and will be illustrated in detail in the next section.  


\textbf{Cross-view image synthesis.}
The cross-view image synthesis task aims to synthesize an image from one viewpoint to another viewpoint. 
This task was first proposed by Regmi and Borji~\cite{regmi2018cross}, where a conditional GAN is used to learn the transformations between the two images.
Since then, different methods have been proposed to improve the performance~\cite{regmi2019cross, tang2019multi, lu2020geometry, shi2021geometry, li2021sat2vid}. 
Recent research shows that cross-view image localization and synthesis tasks can complement each other, improving their performance~\cite{Regmi_2019_ICCV, toker2021coming}. In this work, we use cross-view feature synthesis instead of image synthesis to facilitate cross-view localization performance.

\textbf{Cross-view transformers.}
Overhead view, also known as Bird's eye view (BEV), representation learning~\cite{zhang2014robust, roddick2020predicting, zhou2022cross, chen2022persformer, li2022bevformer, saha2022translating} has also been shown to be useful in many autonomous driving tasks, such as map-view semantic segmentation~\cite{roddick2020predicting, zhou2022cross}, 3D lane line detection~\cite{chen2022persformer}, and 3D object detection~\cite{li2022bevformer}. 
A common strategy is to learn an implicit overhead view embedding and then use the learned embedding as query features to collect ground view features and update the overhead view feature maps. 
Cross-view correspondences are learned implicitly from training. 
Instead of learning cross-view correspondences implicitly, 
this paper proposes a geometry-guided cross-view transformer, which explicitly encodes the geometric correspondences to the learnable transformer. 
This eases the burden of neural networks and significantly reduces training time. 



\section{Boosting 3-DoF Camera Pose Accuracy }
Given a coarse location and rotation estimate of a ground camera, this paper aims to refine this camera pose by ground-to-satellite image matching. 
The coarse camera pose estimates can be provided by city-scale ground-to-satellite image retrieval, VO, SLAM, or imprecise sensors (\eg, consumer-level GPS and compass). 

\subsection{Method Overview}

Humans usually determine a ground camera's pose relative to a satellite image by first mentally hallucinating overhead view appearances of the observed scenes and then comparing them with the satellite map.
Inspired by this, we propose a geometry-guided cross-view transformer for ground-to-overhead view feature synthesis to mimic the ``hallucination'' process. 
Then, a neural pose optimizer is constructed to perform the ``comparison'' behavior. 

The neural optimizer constructed by neural networks can produce accurate and reliable rotation estimations, because a rotation on the input can be magnified on the network output. 
However, a small translation on the optimizer input is likely to be absorbed by high-level deep features inside the optimizer, leading to inaccurately estimated translations. Nonetheless, we have a more effective method for translation estimation.

When the ground camera's orientation is estimated, its location can be computed by a spatial correlation between the synthesized overhead view feature map according to the estimated rotation and the observed satellite feature map. The spatial correlation densely searches every possible location on the satellite image, avoiding local minima problems. We also encode an uncertainty map in the spatial correlation process to exclude impossible camera locations (e.g., areas indicated by buildings and tree canopies). The pixel position corresponding to the maximum correlation result indicates the most likely ground camera's location.

Fig.~\ref{fig:framework} provides an overview of our proposed method. Next, we provide technical details for each component.

\begin{figure}
    \centering
    \includegraphics[width=\linewidth]{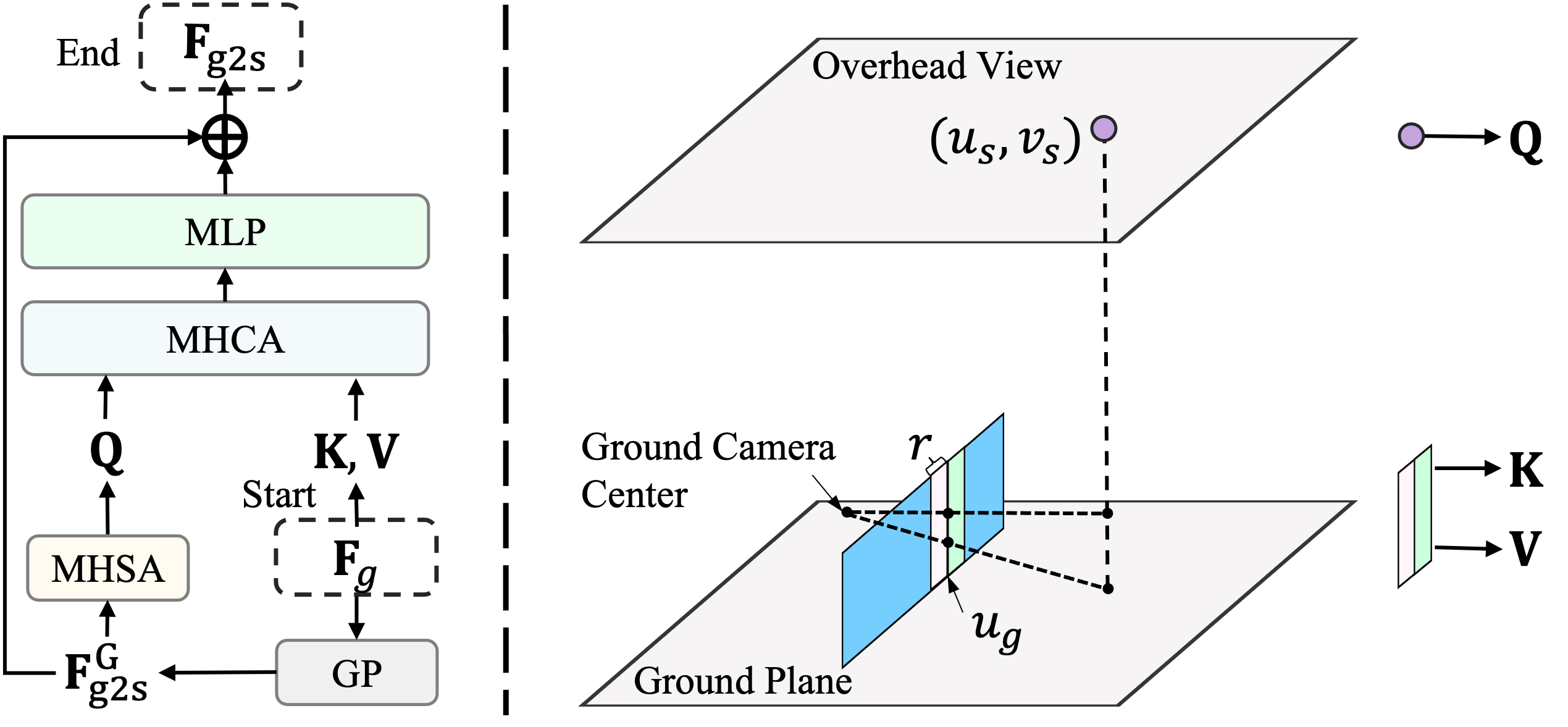}
    \caption{Ground-to-overhead feature synthesis. (Left:) A geometry projection (GP) module is first adopted to project ground image features to an overhead view by exploiting the ground plane homography. Then, a multi-head self-attention (MHSA) module is applied to the GP-projected features $\mathbf{F}_{g2s}^{G}$ to make each element aware of its contextual information. Next, a multi-head cross-attention (MHCA) module is proposed to further collect features from the ground-view images and update the overhead view feature representation. (Right:) For each overhead view feature map pixel ($\mathbf{Q}$), we find its corresponding column in the ground view feature map and the neighboring columns with a radius of $r$, constructing a feature candidate pool ($\mathbf{K, V}$) to be used in MHCA.}
    \label{fig:G2S}
\end{figure}

\subsection{Overhead-view feature synthesis}
Given a ground image, we synthesize an overhead view feature map from the ground view image according to a relative rotation and translation. 
Our overhead-view feature synthesis module combines deterministic geometric correspondences and learnable cross-view transformers. 

\textbf{Geometry correspondences.} 
We use azimuth angle $\theta$ to parameterize the ground camera's relative rotation $\mathbf{R}$ to the satellite image. 
We set the tilt and roll angles as zero since satellite images cannot provide reference for them and they are typically small in autonomous driving scenarios. 
The relative camera translation is parameterized as $\mathbf{t}=[t_x, 0, t_z]^T$, where $t_x$ and $t_z$ denote the translation along the latitude and longitude directions in the geographical coordinates, corresponding to the $v$ and $u$ directions of the satellite image coordinates, respectively. 
The relative translation $t_y$ between a ground and a satellite camera along the vertical direction is infinite. Thus we do not consider it in our task by setting it as zero. 
The mapping from an overhead view pixel $(u_s, v_s)$ to a ground view image pixel $(u_g, v_g)$ is derived as:

\begin{equation}
\begin{bmatrix}
u_g\\ 
v_g
\end{bmatrix} = 
\begin{bmatrix}
f_x  \frac{\left [(v_s - v_s^0) + t_x\right ] \cos \theta - \left [(u_s - u_s^0) + t_z \right ]\sin \theta}{\left [ (v_s - v_s^0) + t_x\right ]\sin \theta + \left [ (u_s - u_s^0) + t_z\right ]\cos \theta} + u_g^0\\ 
f_y \frac{h}{\alpha \left \{ \left [ (v_s - v_s^0) + t_x\right ]\sin \theta + \left [ (u_s - u_s^0) + t_z\right ]\cos \theta \right \}}+ v_g^0
\end{bmatrix}, 
\label{eq:geocorrespondences}
\end{equation}
where $(u_s^0, v_s^0)$ and $(u_g^0, v_g^0)$ denote the overhead view feature map and the ground-view image centers, respectively, $\alpha$ is the ground resolution of the overhead view feature map, $f_x$ an $f_y$ denote the ground camera focal length along $u$ and $v$ directions, respectively, $h$ is the height of pixel $(u_s, v_s)$ above the ground plane. 

It can be seen that the value of $u_g$ can always be determined whenever the height of the corresponding pixel is given or not, and the height only affects the value of $v_g$.

\textbf{Geometry-guided cross-view transformer.} 
Next, we leverage cross-view transformers to handle the ambiguity of $v_g$. 
Commonly designed cross-view transformers~\cite{roddick2020predicting, zhou2022cross, li2022bevformer, chen2022persformer} initialize a latent overhead view embedding that is shared by different overhead view maps. 
The embedding functions as ``query'' features to collect ground-view image features and is learned by statistically data-driven training, enabling the network to collect information from ground-view images automatically.

In the cross-view localization task, the appearance of overhead view features at different geographical locations differs, and these differences are essential to localization accuracy. 
To address this issue, we design a geometry-guided cross-view transformer with a scene-specific ``query'' embedding initialization strategy, as shown in Fig.~\ref{fig:G2S} (left).

First, the geometry projection (GP) module projects the ground-view features to the overhead view by exploiting the ground plane homography.
This establishes authentic cross-view correspondences for scene contents on the ground plane. 
We then apply a multi-head self-attention (MHSA) block to the geometry projected features, making them aware of their context information, especially for the scene contents above the ground plane. 
The output of MHSA is used as our overhead view ``query'' embedding, which is scene-specific. 

We have derived from Eq.~\eqref{eq:geocorrespondences} that the column index $u_g$ of a satellite pixel on the ground view image can always be determined. 
It implies that its corresponding ground-view image pixel lies on a column on the ground-view image. 
When there is a slight tilt or pitch angle change of the ground camera during driving, the value of $u_g$ will be slightly off (the value of $v_g$ remains unknown). 
With this observation, we retrieve the entire column indexed by $u_g$ and its neighbor columns with a radius of $r$ in the ground-view image, as shown in Fig.~\ref{fig:G2S} (right), which is to construct the candidate feature pool (``key'' and ``value'') for this satellite pixel.
We then use the strength of transformers to update the overhead view feature map by multi-head cross-attention (MHCA):

\begin{equation}
     \text{MHCA}\left ( \mathbf{F}_{g2s}^G, \mathbf{F}_g \right ) = \text{Softmax}\left ( {\mathbf{QK}^T}\right ) \mathbf{V},
\end{equation}
\begin{equation}
        \mathbf{Q} = \mathcal{Q} \left( \text{MHSA}\left(\mathbf{F}_{g2s}^{G}\right)\right )  \;
        \mathbf{K} = \mathcal{K} \left( \mathbf{F}_{g}^L \right )   \;
        \mathbf{V} = \mathcal{V} \left( \mathbf{F}_{g}^L \right ) ,
\end{equation}
where $\mathbf{F}_{g}$ denote the ground-view image features map, and $\mathbf{F}_{g}^L$ indicate the retrieved local region for each satellite pixel from $\mathbf{F}_{g}$ illustrated in Fig.~\ref{fig:G2S}, $\mathcal{Q}(\cdot), \mathcal{K}(\cdot), \mathcal{V}(\cdot)$ are linear mapping layers.
We then feed the updated features by MHCA to an MLP layer $\mathcal{M}(\cdot)$ with a skip connection, obtaining the final synthesized overhead view feature map:
\begin{equation}
\mathbf{F}_{g2s} = \mathbf{F}_{g2s}^G + \mathcal{M} \left (\text{MHCA}\left ( \mathbf{F}_{g2s}^G, \mathbf{F}_g \right ) \right )
\end{equation}

For memory efficiency, this cross-view transformer is only applied to the coarsest feature levels (\ie, $\frac{1}{8}$ of the original image size). 
This also guarantees a large receptive field for cross-attention. 
We adopt a decoder to recover finer details of the overhead view map at higher resolutions. 

\subsection{Neural pose optimizer}


The overhead view feature synthesis module has aligned the ground view observations and the satellite image in the same (overhead) domain. We next design a neural pose optimizer to estimate the relative pose between them, especially the relative rotation. 
Our neural optimizer takes input as the differences between the synthesized and observed feature maps $\mathbf{F}_{g2s} - \mathbf{F}_{s}$, where $\mathbf{F}_{s}$ denote the observed satellite image features, and outputs a relative pose update based on the current pose estimation.
It is constructed by two swin transformer layers~\cite{liu2021swin} and two MLP layers.
The swin transformer layers are to increase the global information extraction ability of the neural optimizer to avoid local minima, especially for the estimated rotations. 
Here, we make the optimizer update the translation as well.
The reason is to encourage the optimizer to find a translation, based on which the rotation can be easily and correctly estimated. 
The neural pose optimizer is applied from coarse to finer feature levels and then repeated. 
This allows fine-tuning around a potential global minimum and jumping out of local minima.

\begin{table*}[t!]
\setlength{\abovecaptionskip}{0pt}
\setlength{\belowcaptionskip}{0pt}
\setlength{\tabcolsep}{3pt}
\centering
\footnotesize
\caption{\small Comparison results on KITTI and Ford Multi-AV with aligned orientation. ``*'' indicates fine-grained image retrieval methods.}
\begin{tabular}{c|ccH|ccH|ccH|ccH|ccH|ccH|ccH|ccH}
\toprule
           & \multicolumn{3}{c|}{Lateral}   & \multicolumn{3}{c|}{Longitudinal}  & \multicolumn{3}{c|}{Lateral} & \multicolumn{3}{c|}{Longitudinal}    
            & \multicolumn{3}{c|}{Lateral}   & \multicolumn{3}{c|}{Longitudinal}  & \multicolumn{3}{c|}{Lateral} & \multicolumn{3}{c}{Longitudinal}    \\
           & $d=1$          & $d=3$          & $d=5$          & $d=1$         & $d=3$          & $d=5$         & $d=1$          & $d=3$          & $d=5$          & $d=1$         & $d=3$          & $d=5$  & $d=1$          & $d=3$          & $d=5$          & $d=1$         & $d=3$          & $d=5$         & $d=1$          & $d=3$          & $d=5$          & $d=1$         & $d=3$          & $d=5$      \\\midrule
           & \multicolumn{6}{c|}{KITTI - Test1}                                                                                                                             & \multicolumn{6}{c|}{KITTI - Test2}        
            & \multicolumn{6}{c|}{Ford - Log1}                                                                                                                             & \multicolumn{6}{c}{Ford - Log2}   \\ \midrule
CVM-NET*~\cite{Hu_2018_CVPR}   & 3.87    & 12.38   & 22.22   & 3.81     & 11.16     & 18.27    & 3.87    & 12.38   & 22.22   & 3.81     & 11.16     & 18.27    & 9.62    & 25.33   & 43.48   & 4.10      & 12.29     & 20.05         & 10.33   & 29.17   & 46.63   & 4.11     & 12.80     & 21.17           \\
CVFT*~\cite{shi2020optimal}       & 13.04   & 36.84   & 53.75   & 4.06     & 11.50     & 19.56  & 12.19   & 34.23   & 50.34   & 3.75     & 10.36     & 17.69     & 15.14   & 40.71   & 66.52   & 4.38      & 12.19     & 21.29  & 15.00   & 40.94   & 66.73   & 4.70     & 14.76     & 24.09           \\
SAFA*~\cite{shi2019spatial}       & 13.20   & 36.76   & 54.12   & 4.21     & 12.51     & 20.51 & 13.35   & 36.93   & 55.65   & 4.39     & 11.99     & 19.81      & 11.33   & 31.62   & 51.71   & 4.33      & 13.05     & 21.90  & 15.16   & 42.45   & 69.14   & 4.72     & 13.71     & 22.16       \\
Polar-SAFA*~\cite{shi2019spatial} & 13.44   & 37.11   & 54.57   & 4.96     & 13.12     & 22.69 & 13.76   & 38.42   & 56.75   & 3.92     & 11.27     & 19.00     & 13.38   & 37.19   & 60.95   & 3.86      & 10.81     & 18.57 & 14.57   & 40.41   & 63.86   & 4.35     & 12.74     & 21.79       \\
DSM*~\cite{shi2020looking}        & 13.86   & 38.17   & 55.29   & 4.43     & 11.93     & 20.57  & 13.54   & 37.40   & 54.11   & 4.00     & 11.71     & 19.17   & 12.24   & 33.14   & 52.76   & 4.48      & 12.10     & 19.95  & 12.18   & 33.67   & 53.42   & 4.45     & 12.50     & 20.69         \\
VIGOR*~\cite{zhu2021vigor}      & 14.18   & 38.38   & 56.48   & 4.61     & 13.25     & 21.65  & 13.07   & 36.40   & 55.58   & 4.48     & 12.20     & 20.25     & 11.76   & 35.43   & 51.67   & 6.24      & 17.57     & 27.95  & 19.48   & 62.38   & 81.78   & 5.04     & 15.88     & 25.84        \\
L2LTR*~\cite{yang2021cross}             & 15.19  & 41.93  & 60.16 & 5.33  & 14.37  & 23.72  & 14.65  & 40.36  & 58.99 & 4.69  & 12.40  & 20.95  & 14.19  & 38.10  & 64.00 & 4.71  & 13.29  & 22.14  & 13.52  & 38.07  & 60.88 & 4.45  & 12.96  & 21.01  \\
TansGeo*~\cite{Zhu_2022_CVPR}           & 15.74  & 43.15  & 61.81 & 5.04  & 14.44  & 24.36  & 14.85  & 41.37  & 60.28 & 4.31  & 12.64  & 21.19  & 14.57  & 39.76  & 65.24 & 4.10  & 13.86  & 22.81  & 13.57  & 39.76  & 60.24 & 4.10  & 13.86  & 22.81 \\
LM~\cite{shi2022beyond}         & 52.66   & 82.14   & 87.83   & 4.35     & 14.95     & 24.60  & 37.63   & 69.07   & 78.81   & 4.96     & 14.66     & 25.18      & 64.05   & 83.05   & 89.48   & 11.38     & 22.57     & 28.95  & 47.49   & 79.18   & 88.22   & 5.31     & 15.59     & 25.11          \\
CVML~\cite{xia2022visual}       &  64.27  & 83.12	& 91.12   & 34.77	& 65.04	& 76.36        &  19.54	& 49.83	  & 69.33   & 10.47	   & 25.86	   & 36.81      & 53.00	  &83.67	& 89.62   & 6.19	  & 18.62	  & 28.52  & 22.16	 & 37.54   & 94.42   & 5.31	    & 15.56	    & 25.73             \\
Ours      & \textbf{93.85}	&\textbf{98.44}	&\textbf{99.47}	&\textbf{52.40}	&\textbf{79.75}	&\textbf{84.02}        &  \textbf{55.28}	&\textbf{85.73}	&\textbf{90.18}	&\textbf{17.97}	&\textbf{39.49}	&\textbf{50.72}    & \textbf{80.76}	&\textbf{95.90}	&\textbf{97.10}	&\textbf{28.48}	&\textbf{38.57}	&\textbf{40.38}       &\textbf{78.19}	  & \textbf{89.21}	&\textbf{97.50}	&\textbf{22.86}	&\textbf{40.43}	&\textbf{41.70}        
\\ \bottomrule
\end{tabular}
\label{tab:align}\vspace{-2em}
\end{table*}

\begin{table}[ht!]
\setlength{\abovecaptionskip}{0pt}
\setlength{\belowcaptionskip}{0pt}
\setlength{\tabcolsep}{2pt}
\centering
\footnotesize
\caption{ Comparison results on Oxford RobotCar with aligned orientation. The lower, the better. }
\begin{tabular}{c|cc|cc|cc|cc}
\toprule
     & \multicolumn{2}{c|}{Test1}     & \multicolumn{2}{c|}{Test2}     & \multicolumn{2}{c|}{Test3}     & \multicolumn{2}{c}{Overall}   \\
     & Mean          & Median        & Mean          & Median        & Mean          & Median        & Mean          & Median        \\ \midrule
CVML~\cite{xia2022visual} & \textbf{1.66} & 1.29          & \textbf{2.28} & 1.55          & \textbf{2.26} & 1.51          & \textbf{2.07} & 1.44          \\
Ours & 2.40          & \textbf{0.91} & 3.10          & \textbf{1.13} & 2.86          & \textbf{1.05} & 2.79          & \textbf{1.02} \\ \bottomrule
\end{tabular}
\label{tab:oxford_align}\vspace{-1em}
\end{table}

\begin{table}[t!]
\setlength{\abovecaptionskip}{0pt}
\setlength{\belowcaptionskip}{0pt}
\setlength{\tabcolsep}{11pt}
\centering
\footnotesize
\caption{\small Comparison results on VIGOR with aligned orientation.}
\begin{tabular}{c|cc|cc}
\hline
\multirow{2}{*}{Algorithms} & \multicolumn{2}{c|}{Same-area} & \multicolumn{2}{c}{Cross-area} \\
                            & Mean          & Median        & Mean           & Median        \\ \midrule
CVML~\cite{xia2022visual}                         & 6.94          & 3.64          & 9.05           & 5.14          \\
SliceMatch~\cite{lentsch2022slicematch}                  & 5.18          & 2.58          & 5.53           & 2.55          \\
\textbf{Ours}               & \textbf{4.12} & \textbf{1.34} & \textbf{5.16}  & \textbf{1.40} \\ \hline
\end{tabular}
\label{tab:VIGOR}\vspace{-1em}
\end{table}

\subsection{Uncertainty-guided spatial correlation}
After the ground camera's orientation has been estimated, we re-synthesize an overhead view feature map from the ground view observation according to the estimated orientation and zero translation. 
The zero translation here is to make the ground camera's location correspond to the center of the synthesized overhead view feature map. 
Then, we compute the relative translation between the query camera location and the satellite image center by a spatial correlation between the synthesized overhead-view features and the observed satellite image features. This is shown in the numerator of Eq.~\eqref{eq:uncer_corr}. We also consider that the satellite semantics themselves encode a likelihood of the ground vehicles' location, such that a car is likely on a road but unlikely on top of buildings. To account for this, we estimate an uncertainty map from the satellite image features using a set of CNNs and develop an uncertainty-guided similarity-matching scheme:

\begin{equation}
      \mathbf{P} = {\left (\mathbf{F}_{s} \star  \mathbf{F}_{g2s} \right )(u_s, v_s)}/{\mathbf{U}(u_s, v_s)}, 
      \label{eq:uncer_corr}
\end{equation}
where $\star$ denotes the normalized cross-correlation, $\mathbf{U}$ is the uncertainty map estimated from satellite semantics (features). 
The value of the uncertainty is within the range of $(0, 1)$. The higher the value, the lower the possibility of the vehicle being at the corresponding location.

The correlation results, $\mathbf{P}$, indicate the final computed possibility of the query camera being at each of the satellite image pixels. 
A higher value implies a large probability of the query camera being at the corresponding location. 
We regard the pixel location with the highest correlation value as the query camera location.


\subsection{Training objective}

We use ground truth pose as supervision for the neural optimizer output:
\begin{equation}
    \mathcal{L}_1 = \sum_n \sum_l (\|{\theta}_n^l - \theta^*\|_1 + \|{t_x}_n^l - {t_x}^*\|_1)+ \|{t_z}_n^l - {t_z}^*\|_1), 
\end{equation}
where $l=[1, 2, 3]$ and $n = [1, 2]$ index the feature levels and the number of iterations, respectively, ${\theta}_n^l, {t_x}_n^l, {t_z}_n^l$ denote the predicted camera pose (rotation and translation) by the neural optimizer at feature level $l$ and iteration number $n$, $\theta^*, {t_x}^*, {t_z}^*$ represent the ground truth (GT) camera pose. 
The predicted translation by the neural optimizer is not taken as our final output, but we still apply supervision for them here.  
This is to encourage the gradients of trainable parameters in the neural optimizer to be updated smoothly and in correct directions.

We apply a triplet loss to the spatial correlation results for translation estimation.
We aim to maximize the possibility at GT (positive) camera location while minimizing that at other (negative) locations: 
\begin{equation}
   \mathcal{L}_2 = \frac{1}{\left | \mathbf{P} \right |}\sum_{(u_s, v_s)} log(1 + e^{\gamma \left ( \mathbf{P}(u_s^*, v_s^*) - \mathbf{P}(u_s, v_s) \right )}), 
   \label{eq:triplet}
\end{equation}
where $(u_s^*, v_s^*)$ indicates the pixel corresponding to the GT camera location, $(u_s, v_s)$ denotes other pixels, $\left | \mathbf{P} \right |$ is the total number of pixels in $\mathbf{P}$, and $\gamma$ is set to $10$.
We should note that we do not provide explicit supervision for the uncertainty maps in Eq.~\eqref{eq:uncer_corr}. Rather, they are learned statistically and implicitly from the translational pose training loss represented by Eq.~\eqref{eq:triplet}. 

We follow Kendall and Roberto~\cite{kendall2017geometric} to balance the two training loss items:

\begin{equation}
    \mathcal{L} = \mathcal{L}_1 e^{-\lambda_1} + \lambda_1 + \mathcal{L}_2 e^{-\lambda_2} + \lambda_2 
\end{equation}
where $\lambda_1$ is initialized as $-5$ and $\lambda_2$ as $-3$. They are learned and adjusted dynamically during training.

\section{Experiments}


\textbf{Dataset and evaluation metrics.}
We evaluate the performance of our method and compare it with state-of-the-art on several datasets, including the cross-view KITTI~\cite{geiger2013vision, shi2022beyond} and Ford Multi-AV~\cite{agarwal2020ford, shi2022beyond} datasets, the Oxford RobotCar dataset~\cite{maddern20171, xia2022visual} as well as the VIGOR dataset~\cite{zhu2021vigor}.

The cross-view \textbf{KITTI} dataset consists of one training set and two test sets. Test1 contains images sampled from the same region as the training set, while Test2 contains images from a different region.
The cross-view \textbf{Ford Multi-AV} dataset~\cite{agarwal2020ford} includes six subsets (Log1-Log6), each captured on two different dates along different trajectories. Images from one date are used for training, while those from the other are used for testing. We follow Shi and Li~\cite{shi2022beyond} for the evaluation of these two datasets. 
Specifically, a query image is considered correctly localized in a direction if its estimated translation along that direction is within $d$ meters of the ground truth translation. The percentage of correctly localized query images in that direction is recorded. We also report the percentage of images with correctly estimated orientation, when the estimated orientation of the query image is within $\theta$ degrees of its ground truth orientation. 
We present results for the first two logs of the Ford dataset in the main paper and include results for the remaining logs in the supplementary material.

The cross-view \textbf{Oxford RobotCar} dataset includes one training set, one validation set, and three test sets. The test sets consist of images from three traversals captured at different dates than those in the training set. The \textbf{VIGOR} dataset contans cross-view images from four cities in the USA, Chicago, New York, San Francisco, and Seattle. The dataset contains two evaluation scenarios: same-area and cross-area evaluation. We use the rectified labels by SliceMatch~\cite{lentsch2022slicematch} and following its evaluation protocol in only using the fully positive satellite images. Both the Oxford RobotCar and the VIGOR datasets assumes orientation alignment and does not provide a test set with unknown orientations. Therefore, we only test location estimation with aligned orientation. We follow the evaluation protocol outlined in its original paper~\cite{xia2022visual}, reporting mean and median distances between predicted and ground truth locations.
We perform joint location and orientation estimation on the KITTI and Ford Multi-AV datasets, as they provide official test sets for this challenging scenario.



\begin{table*}[t!]
\setlength{\abovecaptionskip}{0pt}
\setlength{\belowcaptionskip}{0pt}
\setlength{\tabcolsep}{2.5pt}
\centering
\footnotesize
\caption{\small Performance comparison on KITTI and Ford Multi-AV with $20^\circ$ orientation noise. }
\begin{tabular}{c|ccc|ccc|ccc|ccc|ccc|ccc}
\toprule
           & \multicolumn{3}{c|}{Lateral}                          & \multicolumn{3}{c|}{Longitudinal}                         & \multicolumn{3}{c|}{Azimuth}                      & \multicolumn{3}{c|}{Lateral}                          & \multicolumn{3}{c|}{Longitudinal}                         & \multicolumn{3}{c}{Azimuth}                      \\
           & $d=1$          & $d=3$          & $d=5$          & $d=1$         & $d=3$          & $d=5$          & $\theta=1$     & $\theta=3$     & $\theta=5$     & $d=1$          & $d=3$          & $d=5$          & $d=1$         & $d=3$          & $d=5$          & $\theta=1$     & $\theta=3$     & $\theta=5$     \\\midrule
           & \multicolumn{9}{c|}{KITTI - Test1}                                                                                                                             & \multicolumn{9}{c}{KITTI - Test2}                                                                                                                             \\\midrule
DSM*~\cite{shi2020looking}        & 10.12          & 30.67          & 48.24          & 4.08          & 12.01          & 20.14          & 3.58           & 13.81          & 24.44          & 10.77          & 31.37          & 48.24          & 3.87          & 11.73          & 19.50          & 3.53           & 14.09          & 23.95          \\ 
LM~\cite{shi2022beyond}      & {35.54} & {70.77} & {80.36} & {5.22} & {15.88} & {26.13} & {19.64} & {51.76} & {71.72} & {27.82} & {59.79} & {72.89} & {5.75} & {16.36} & {26.48} & {18.42} & {49.72} & {71.00}\\
SliceMatch~\cite{lentsch2022slicematch} & 49.49 & -- & 98.52 & 15.19 & -- & 57.35 & 13.41 & -- & 64.17 & 32.43 & -- & 86.44 & 8.30 & -- & 35.57 & 46.82 & -- & 46.82 \\
{Ours}       & \textbf{76.44}	&\textbf{96.34}	&\textbf{98.89}	&\textbf{23.54}	&\textbf{50.57}	&\textbf{62.18}	&\textbf{99.10}	&\textbf{100.00}	&\textbf{100.00} &\textbf{57.72}	&\textbf{86.77}	&\textbf{91.16}	&\textbf{14.15}	&\textbf{34.59}	&\textbf{45.00}	&\textbf{98.98}	&\textbf{100.00}	&\textbf{100.00}\\ \bottomrule
           & \multicolumn{9}{c|}{Ford - Log1}                                                                                                                             & \multicolumn{9}{c}{Ford - Log2}                                                                                                                             \\ \midrule
DSM*~\cite{shi2020looking}        & 12.00          & 35.29          & 53.67          & 4.33          & 12.48          & 21.43          & 3.52           & 13.33          & 23.67          & 8.45           & 24.85          & 37.64          & 3.94          & 12.24          & 21.41          & 2.23          & 7.67           & 13.42          \\
LM~\cite{shi2022beyond}       & {46.10} & {70.38} & {72.90} & 5.29          & {16.38} & {26.90} & {44.14} & {72.67} & {80.19} & {31.20} & {66.46} & {78.27} & 4.80          & 15.27          & 25.76          & {9.74} & {30.83} & {51.62}
\\
{Ours}       & \textbf{70.00}	&\textbf{93.10}	&\textbf{95.52}	&\textbf{17.10}	&\textbf{27.95}	&\textbf{29.76}	&\textbf{59.38}	&\textbf{90.57}	&\textbf{95.24} & \textbf{60.26}	& \textbf{76.58}	& \textbf{95.12}	& \textbf{20.77}	& \textbf{37.67}	& \textbf{39.39}	& \textbf{62.68}	& \textbf{93.78}	& \textbf{98.77 }\\ \bottomrule
\end{tabular}
\label{tab:unknown}
\end{table*}

\begin{figure*}[ht]
\setlength{\abovecaptionskip}{0pt}
\setlength{\belowcaptionskip}{0pt}
    \centering
    \begin{minipage}{0.9\linewidth}
    \centering
        \begin{minipage}{0.32\linewidth}
        \includegraphics[width=\linewidth, height=0.5\linewidth]{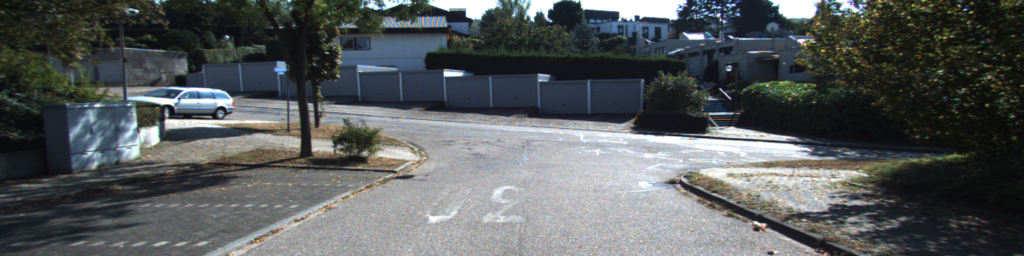}
        \includegraphics[width=\linewidth, height=0.5\linewidth]{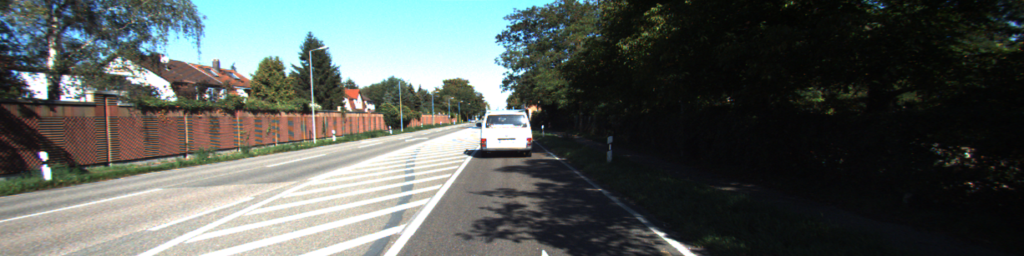}
        \centerline{\small Query Image}
    \end{minipage}
    \begin{minipage}{0.16\linewidth}
        \adjincludegraphics[width=\linewidth,trim={{0.23\width} {0.23\width} {0.23\width} {0.23\width}},clip]{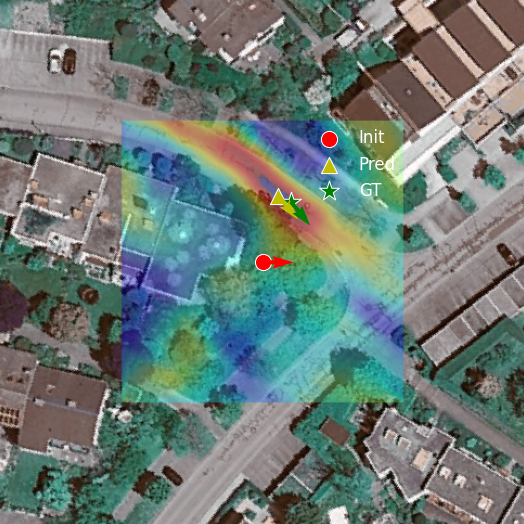}
        \adjincludegraphics[width=\linewidth,trim={{0.23\width} {0.23\width} {0.23\width} {0.23\width}},clip]{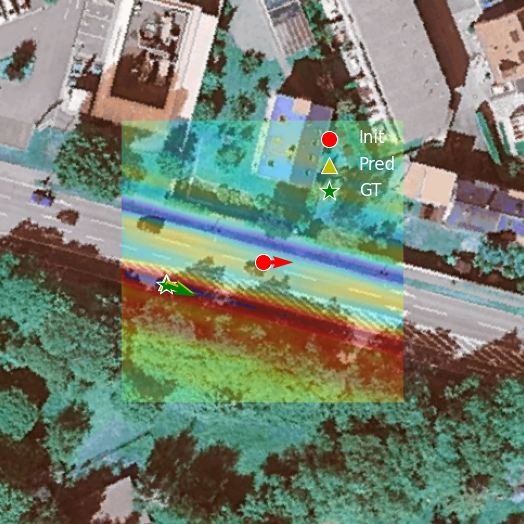}
        \centerline{\small Prediction}
    \end{minipage}
    \hspace{0.5em}
    \begin{minipage}{0.32\linewidth}
        \includegraphics[width=\linewidth, height=0.5\linewidth]{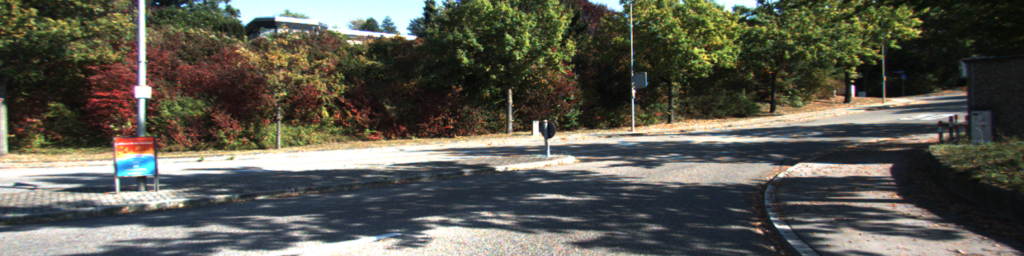}
        \includegraphics[width=\linewidth, height=0.5\linewidth]{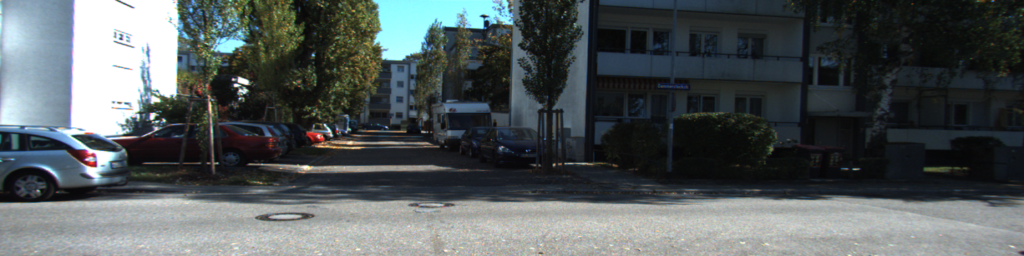}
        \centerline{\small Query Image}
    \end{minipage}
    \begin{minipage}{0.16\linewidth}
        \adjincludegraphics[width=\linewidth,trim={{0.23\width} {0.23\width} {0.23\width} {0.23\width}},clip]{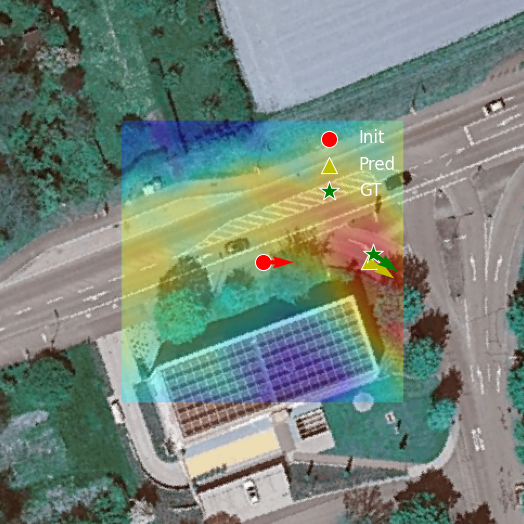}
        \adjincludegraphics[width=\linewidth,trim={{0.23\width} {0.23\width} {0.23\width} {0.23\width}},clip]{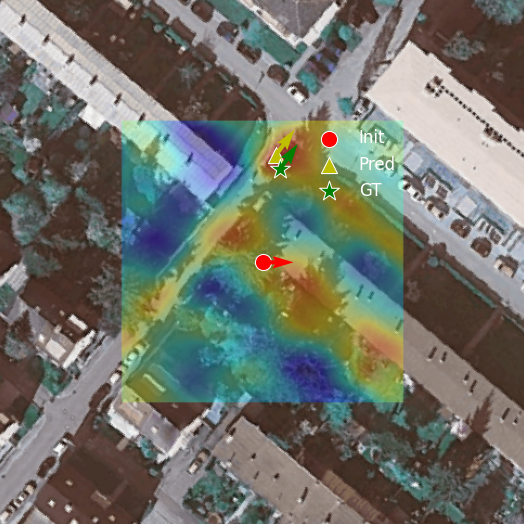}
        \centerline{\small Prediction}
    \end{minipage}
    \end{minipage}
    \caption{Localization results visualization. The correlation results (vehicle location possibility maps) overlay the satellite images}. 
    \label{fig:visualize}\vspace{-2em}
\end{figure*}

\textbf{Implementation details.}
Our network backbone uses a U-Net architecture with the encoder being a VGG16~\cite{Simonyan2014VeryDC} pretrained on ImageNet~\cite{deng2009imagenet}. The decoder weights are randomly initialized. The satellite branch has two output heads: one for satellite features and the other for uncertainty maps. The satellite image resolution for all datasets is set to $512\times512$, following the original settings~\cite{shi2022beyond, xia2022visual}. The ground image resolution is $256\times1024$ for the cross-view KITTI and Ford Multi-AV datasets, and $154\times231$ for the cross-view Oxford RobotCar dataset. The local region radius in MHCA is set to 1. The learning rate starts from $10^{-4}$ and gradually decreases to $10^{-5}$. The network is trained for five epochs with a batch size of 3. Using an RTX 3090 GPU, the inference time for each image is 280ms, significantly faster than Deep LM~\cite{shi2022beyond}, which requires 500ms. Code is available at \url{https://github.com/shiyujiao/Boosting3DoFAccuracy.git}.

\subsection{Comparison with the state-of-the-art}

We compare our method with the state-of-the-art, including fine-grained image retrieval~\cite{Hu_2018_CVPR, shi2020optimal, shi2019spatial, shi2020looking, zhu2021vigor}, deep LM optimization~\cite{shi2022beyond} and cross-view metric learning (CVML)~\cite{xia2022visual}. 
For the fine-grained image retrieval, we split the satellite image into $N \times N$ small patches and retrieve the most similar patch. The retrieved patch center is regarded as the query camera location. 
Results are from paper~\cite{shi2022beyond} or re-evaluated by the author-provided models and codes. 
All experiment settings follow previous works~\cite{shi2022beyond, xia2022visual}.

\begin{table*}[t!]
\setlength{\abovecaptionskip}{0pt}
\setlength{\belowcaptionskip}{0pt}
\setlength{\tabcolsep}{4pt}
\centering
\footnotesize
\caption{\small Ablation study results on KITTI with $20^\circ$ orientation noise.}
\begin{tabular}{c|c|ccH|ccH|ccH|ccH|ccH|ccH}
\toprule
      &     & \multicolumn{3}{c|}{Lateral}                          & \multicolumn{3}{c|}{Longitudinal}                         & \multicolumn{3}{c|}{Azimuth}                      & \multicolumn{3}{c|}{Lateral}                          & \multicolumn{3}{c|}{Longitudinal}                         & \multicolumn{3}{c}{Azimuth}                      \\
       &    & $d=1$          & $d=3$          & $d=5$          & $d=1$         & $d=3$          & $d=5$          & $\theta=1$     & $\theta=3$     & $\theta=5$     & $d=1$          & $d=3$          & $d=5$          & $d=1$         & $d=3$          & $d=5$          & $\theta=1$     & $\theta=3$     & $\theta=5$     \\\midrule
       &   & \multicolumn{9}{c|}{Test1}                                                                                                                             & \multicolumn{9}{c}{Test2}                                                                                                                             \\\midrule
\multirow{3}{*}{\begin{tabular}[c]{@{}c@{}}G2S Synthesis \\ Alternatives\end{tabular}}      & Geometry Projection        & 75.96          & 95.68          & 98.14          & 22.55          & 46.67          & 55.79          & 97.59          & \textbf{100.00} & \textbf{100.00} & 54.80          & 85.68          & 90.03          & 14.00          & 30.99          & 40.18          & 98.34          & 100.00          & \textbf{100.00} \\
                                                                                  & Persformer~\cite{chen2022persformer}      & 68.04          & 93.85          & 97.32          & 5.51           & 15.77          & 26.50          & 97.98          & 99.97           & \textbf{100.00} & 38.86          & 73.56          & 83.31          & 3.71           & 11.77          & 20.11          & 98.42          & 100.00          & \textbf{100.00} \\
                                                                                  & Ours & \textbf{76.44} & \textbf{96.34} & \textbf{98.89} & \textbf{23.54} & \textbf{50.57} & \textbf{62.18} & \textbf{99.10} & \textbf{100.00} & \textbf{100.00} & \textbf{57.72} & \textbf{86.77} & \textbf{91.16} & \textbf{14.15} & \textbf{34.59} & \textbf{45.00} & \textbf{98.98} & \textbf{100.00} & \textbf{100.00} \\ \midrule
\multirow{2}{*}{Uncertainty}                                                      & Without         & 70.29          & 94.30          & 97.35          & 18.29          & 40.39          & 50.01          & 84.44          & 99.84           & \textbf{100.00} & 54.53          & 85.44          & 89.88          & 12.50          & 29.65          & 39.59          & 84.16          & 99.80           & \textbf{100.00} \\
                                                                                  & With (Ours)     & \textbf{76.44} & \textbf{96.34} & \textbf{98.89} & \textbf{23.54} & \textbf{50.57} & \textbf{62.18} & \textbf{99.10} & \textbf{100.00} & \textbf{100.00} & \textbf{57.72} & \textbf{86.77} & \textbf{91.16} & \textbf{14.15} & \textbf{34.59} & \textbf{45.00} & \textbf{98.98} & \textbf{100.00} & \textbf{100.00}
\\\bottomrule
\end{tabular}
\label{tab:ablation} \vspace{-2em}
\end{table*}

\textbf{Location estimation with given orientation.}
\label{subsec:orien_align}
Since most of the existing cross-view localization methods have an assumption of aligned orientation and only target location estimation, we first compare the performance of our method with them on localizing orientation-aligned images. 
The results on KITTI and Ford Multi-AV datasets are presented in Tab.~\ref{tab:align}. 
It can be seen that all fine-grained image retrieval-based methods achieve inferior results.
This is because the images in the fine-grained database are too similar to disambiguate. 
Benefiting from the specific design for the accurate cross-view localization task, the Deep LM~\cite{shi2022beyond} and CVML~\cite{xia2022visual} significantly improve the results. 
Nonetheless, our method increases the performance considerably, \eg from $64.27\%$ to $93.85\%$ on Test1 of KITTI.

The comparison results on the Oxford RobotCar and VIGOR datasets are presented in Tab.~\ref{tab:oxford_align} and Tab.~\ref{tab:VIGOR}. 
We also compare with SliceMatch~\cite{lentsch2022slicematch}, a contemporary work with our submission, on the VIGOR dataset. The results are from its original paper. 
The comparison reveals that our method achieves the best performance on most scenarios. 
Our method achieves a better median error while a worse mean error than CVML on the Oxford RobotCar dataset, with a similar pattern mirrored on the VIGOR dataset, where our improvement on the median error over CVML is more significant than on the mean error. 
The reason is that our method utilizes similarity matching instead of network regression for location estimation. 

When the scene is diverse and no symmetry exists, the location computed by similarity matching is closer to the GT value than regressed by networks, leading to a smaller median error. 
However, in rare cases where scene symmetry exists at ``different and distant'' locations, the location computed by similarity matching can be far from the GT value. This increases the mean error of our method. 
In contrast, the network regression-based method ({CVML}) tends to resemble the GT values, resulting in a smaller mean error, especially in cases when training and testing images are from the same area (RobotCar).
The {VIGOR} dataset contains images from four cities (Vs. one city in RobotCar), and the number of testing images in {VIGOR} is around $10\times$ larger than those in RobotCar. 
Thus, the {VIGOR} dataset is more diverse, and our method achieves consistently better results than {CVML}. 
More importantly, our improvement over {CVML} is more considerable on the cross-area evaluation than the same-area evaluation on {VIGOR}, and our generalization from Test1 to Test2 on the KITTI dataset is also better than CVML.

\textbf{Joint location and orientation estimation.}
Next, we evaluate the performance of different methods on joint location and orientation estimation. 
Among the comparison algorithms, only DSM~\cite{shi2020looking} and Deep LM~\cite{shi2022beyond} are designed for this purpose. 
The results are presented in Tab.~\ref{tab:unknown}. 
It can be seen that our method achieves the best performance among the comparison algorithms on all the test sets. 
Remarkably, we obtain an exceptional $99\%$ likelihood for restricting the estimated rotation to be $1^\circ$ to its GT value for both test sets on KITTI, improving from around $19.64\%$ on Test1 and $46.82\%$ on Test2. 
This should be attributed to the powerfulness of our neural optimizer on rotation estimation. 
Benefiting from the high rotation estimation accuracy and the dense search scheme, we nearly double the results on lateral pose estimation with $d=1$ for almost all the test sets.
Fig.~\ref{fig:visualize} shows some examples of our localization results.


\subsection{Model analysis}


\begin{figure}[t]
\setlength{\abovecaptionskip}{0pt}
\setlength{\belowcaptionskip}{0pt}
    \centering
    \begin{minipage}{\linewidth}
        \includegraphics[width=0.24\linewidth]{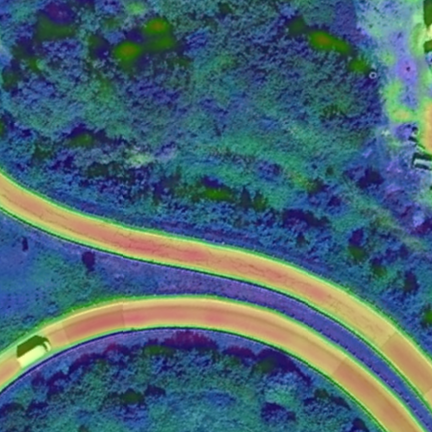}
    \includegraphics[width=0.24\linewidth]{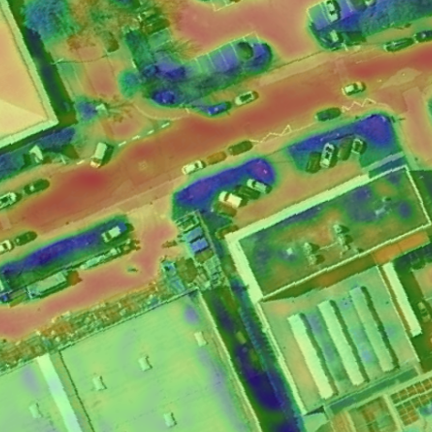}
    \includegraphics[width=0.24\linewidth]{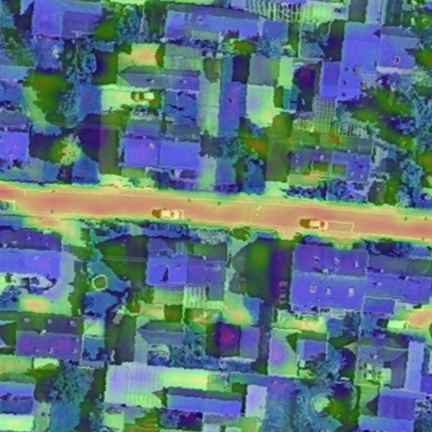}
    \includegraphics[width=0.24\linewidth]{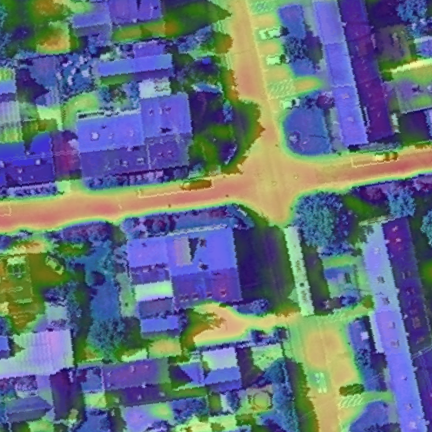}\\
    \includegraphics[width=0.24\linewidth]{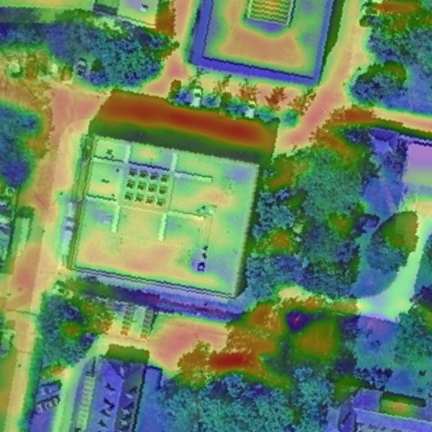}
    \includegraphics[width=0.24\linewidth]{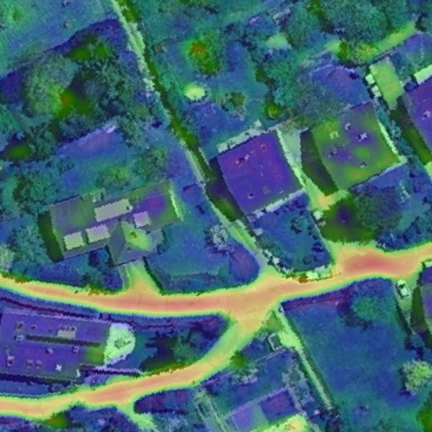}
    \includegraphics[width=0.24\linewidth]{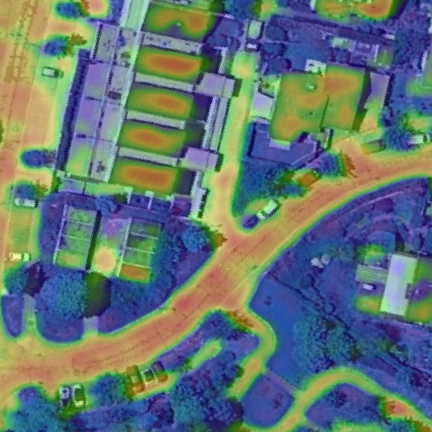}
    \includegraphics[width=0.24\linewidth]{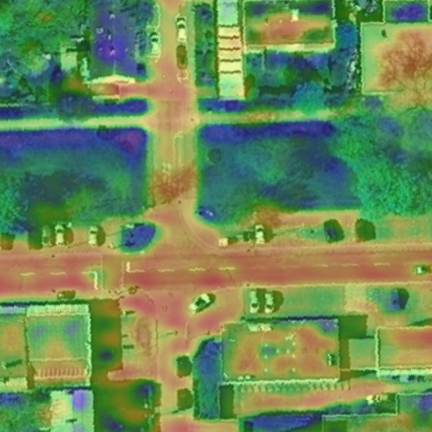}
    \end{minipage}
    \caption{Learned inverse uncertainty maps from satellite features, where road regions are highlighted with small uncertainty. }
    \label{fig:uncertainty} \vspace{-1em}
\end{figure}

\textbf{G2S feature synthesis alternatives.}
We first compare different ground-to-satellite feature synthesis alternatives, including pure geometry projection~\cite{shi2022beyond} and Persformer~\cite{chen2022persformer} (a geometry-guided deformable cross-view transformer). 
Since our method leverages the merits of both conventional geometry and learnable cross-view transformers, our method achieves the best performance compared to the different alternatives, as indicated in the first part of Tab.~\ref{tab:ablation}.

\textbf{Effectiveness of uncertainty maps.}
Next, we conduct experiments to demonstrate the effectiveness of uncertainty maps estimated from satellite images. 
Fig.~\ref{fig:uncertainty} shows some examples of the inverse uncertainty (confidence) maps with road regions highlighted. These regions represent the likelihood of the presence of a vehicle.
The last two rows in Tab.~\ref{tab:ablation} present the performance of our method with or without uncertainty maps in translation estimation. 
It can be seen that even without the uncertainty maps, our method can still achieve promising localization results. 
With the uncertainty maps, the performance is further improved.

Due to space limits, we discuss the comparison between ground-to-satellite projection and satellite-to-ground projection, the sensitivity of our method to different initial poses, the iteration number choice of the neural pose optimizer, and additional ablation study results in the supplementary material. 


\section{Discussion and Conclusion}
Various methods can provide a coarse estimation of a camera's pose at ground level, such as city-scale ground-to-satellite image retrieval, VO, SLAM, or sensors (noisy GPS, compass, wheel encoder, etc.). Given a coarse rotation and translation estimate, this paper has presented a new framework to improve the ground camera's pose accuracy by ground-to-satellite image matching. This task has various applications, such as reducing costs in autonomous driving and field robotics by using a cheaper GPS device. It can also function as a new loop closure method for SLAM and VO techniques.

Our approach consists of a geometry-guided cross-view transformer for ground-to-overhead feature synthesis, a neural pose optimizer for rotation estimation, and an uncertainty-guided spatial correlation for translation estimation. It significantly outperforms the current state-of-the-art in various challenging localization scenarios.


{\small
\bibliographystyle{ieee_fullname}
\bibliography{egbib}
}

\newpage
\onecolumn
\appendix
\appendixpage

\section{Performance on Log3-6 in Ford Multi-AV}
We present the performance of deep LM~[{\color{green}29}], CVML~[{\color{green}43}] and our method on remaining logs in the cross-view Ford multi-AV dataset with aligned-orientation in Tab.~\ref{tab:ford_align}. 
It can be seen that our method achieves significantly better performance in most cases. 
Since CVML cannot estimate orientation, we only compare with deep LM on joint location and orientation estimation. 
The results are provided in Tab.~\ref{tab:ford_unknown}.
Our method achieves consistently better performance. 

\begin{table*}[ht]
\setlength{\abovecaptionskip}{0pt}
\setlength{\belowcaptionskip}{0pt}
\setlength{\tabcolsep}{3pt}
\centering
\footnotesize
\caption{\small Performance comparison on remaining logs of Ford Multi-AV with aligned-orientation.}
\begin{tabular}{c|ccH|ccH|ccH|ccH|ccH|ccH|ccH|ccH}
\toprule
           & \multicolumn{3}{c|}{Lateral}   & \multicolumn{3}{c|}{Longitudinal}  & \multicolumn{3}{c|}{Lateral} & \multicolumn{3}{c|}{Longitudinal}    
            & \multicolumn{3}{c|}{Lateral}   & \multicolumn{3}{c|}{Longitudinal}  & \multicolumn{3}{c|}{Lateral} & \multicolumn{3}{c}{Longitudinal}    \\
           & $d=1$          & $d=3$          & $d=5$          & $d=1$         & $d=3$          & $d=5$         & $d=1$          & $d=3$          & $d=5$          & $d=1$         & $d=3$          & $d=5$  & $d=1$          & $d=3$          & $d=5$          & $d=1$         & $d=3$          & $d=5$         & $d=1$          & $d=3$          & $d=5$          & $d=1$         & $d=3$          & $d=5$      \\\midrule
           & \multicolumn{6}{c|}{Ford - Log3}                                                                                                                             & \multicolumn{6}{c|}{Ford - Log4}        
            & \multicolumn{6}{c|}{Ford - Log5}                                                                                                                             & \multicolumn{6}{c}{Ford - Log6}   \\ \midrule
LM~[{\color{green}29}]   & 20.40          & 43.27          & 64.73          & 5.13          & 14.80          & 23.33          & 41.81                              & 79.86                              & 87.41                              & 6.61                               & 19.14                              & 30.73                              & 23.86          & 59.11          & 72.51          & 4.37           & 18.86          & 30.77          & 15.6          & 46.3          & 63.4          & 5.1           & 14.9          & 26.4          \\
CVML~[{\color{green}43}] & 10.07          & 42.27          & \textbf{65.07} & 4.33          & 15.4           & 26.067         & 44.94                              & 80.23                              & 98.66                              & 13.05                              & 31.36                              & 46.23                              & 16.77          & 49.31          & 64.23          & 5.57           & 17.06          & 28.54          & \textbf{29.6} & 66.2          & 79.1          & 4.8           & 16.2          & 25.9          \\
Ours & \textbf{36.67} & \textbf{47.47} & 63.20          & \textbf{6.67} & \textbf{19.20} & \textbf{27.67} & \textbf{88.38} & \textbf{99.12} & \textbf{99.43} & \textbf{34.72} & \textbf{63.23} & \textbf{72.80} & \textbf{26.69} & \textbf{71.86} & \textbf{83.23} & \textbf{13.77} & \textbf{29.14} & \textbf{34.86} & 27.3          & \textbf{75.9} & \textbf{83.6} & \textbf{11.9} & \textbf{24.1} & \textbf{31.7}
\\ \bottomrule
\end{tabular}
\label{tab:ford_align}
\end{table*}

\begin{table*}[ht]
\setlength{\abovecaptionskip}{0pt}
\setlength{\belowcaptionskip}{0pt}
\setlength{\tabcolsep}{2.5pt}
\centering
\footnotesize
\caption{\small Performance comparison on remaining logs of Ford Multi-AV with $20^\circ$ orientation noise. }
\begin{tabular}{c|ccc|ccc|ccc|ccc|ccc|ccc}
\toprule
           & \multicolumn{3}{c|}{Lateral}                          & \multicolumn{3}{c|}{Longitudinal}                         & \multicolumn{3}{c|}{Azimuth}                      & \multicolumn{3}{c|}{Lateral}                          & \multicolumn{3}{c|}{Longitudinal}                         & \multicolumn{3}{c}{Azimuth}                      \\
           & $d=1$          & $d=3$          & $d=5$          & $d=1$         & $d=3$          & $d=5$          & $\theta=1$     & $\theta=3$     & $\theta=5$     & $d=1$          & $d=3$          & $d=5$          & $d=1$         & $d=3$          & $d=5$          & $\theta=1$     & $\theta=3$     & $\theta=5$     \\\midrule
           & \multicolumn{9}{c|}{Ford - Log3}                                                                                                                             & \multicolumn{9}{c}{Ford - Log4}                                                                                                                             \\\midrule

LM~[{\color{green}29}]   & 11.40 & 34.00          & 58.13          & 4.47           & 13.13          & 22.47          & 8.93           & 29.73          & 48.80          & 29.96          & 66.28          & 74.88          & 4.96           & 15.52          & 25.92          & 14.33          & 43.69          & 67.45          \\
Ours & \textbf{29.40} & \textbf{57.87} & \textbf{63.00} & \textbf{6.07}  & \textbf{17.53} & \textbf{25.00} & \textbf{26.67} & \textbf{61.93} & \textbf{90.20} & \textbf{79.58} & \textbf{97.78} & \textbf{98.72} & \textbf{15.98} & \textbf{39.50} & \textbf{53.77} & \textbf{64.17} & \textbf{97.04} & \textbf{99.49} \\ \midrule

& \multicolumn{9}{c|}{Ford - Log5}                                                                                                                             & \multicolumn{9}{c}{Ford - Log6}                                                                                                                             \\ \midrule
LM~[{\color{green}29}]   & 15.26 & 54.60          & 76.71          & 6.23           & 19.89          & 32.34          & 17.74          & 47.60          & 67.74          & 20.20          & 45.20          & 59.00          & 3.90           & 14.30          & 24.50          & 10.80          & 31.80          & 52.50          \\
Ours & \textbf{22.66} & \textbf{70.14} & \textbf{90.71} & \textbf{15.43} & \textbf{31.80} & \textbf{38.74} & \textbf{63.71} & \textbf{88.83} & \textbf{93.71} & \textbf{30.80} & \textbf{68.70} & \textbf{78.90} & \textbf{9.70}  & \textbf{23.70} & \textbf{29.10} & \textbf{55.10} & \textbf{93.00} & \textbf{98.60} \\ \bottomrule
\end{tabular}
\label{tab:ford_unknown}
\end{table*}

\section{Why Ground-to-Satellite Synthesis}

The satellite-to-ground (S2G) synthesis has been demonstrated to be superior to ground-to-satellite (G2S) synthesis in joint location and orientation optimization by Shi and Li~[{\color{green}29}]. 
This section demonstrates that G2S synthesis is more essential than S2G synthesis for pure orientation optimization, especially when the orientation ambiguity is significant.

Particularly, S2G synthesis leads to significant information loss, as a ground-view image only corresponds to a portion of the satellite image. 
Consider two ground-view images at the same location. Only when their orientation difference is slight can their overlap be sufficient to predict their relative orientation. 
Their overlap will be significantly reduced when a more considerable orientation difference occurs, making orientation estimation intractable. 
In contrast, G2S synthesis reserves all information from the ground image, and its comparison counterpart (the satellite image) contains the entire scene content. 
Hence, a larger possibility of correct orientation estimation can be guaranteed.

We compare G2S and S2G synthesis for orientation estimation under the proposed pipeline. 
The results are provided in Tab.~\ref{tab:G2S_S2G}. 
Note that the location estimation in this comparison is kept the same as in our proposed method.
It can be seen that the rotation estimation accuracy by G2S synthesis is always better than that of S2G synthesis. 
When the rotation noise becomes significant, the performance discrepancy becomes larger. 
Benefiting from the high rotation estimation performance by G2S synthesis, the location estimation performance under the same pipeline is also better.

The other advantage of G2S over S2G synthesis is its superior efficiency in location estimation when the orientation has already been estimated. G2S synthesis retains all the information from the ground view, and the synthesized features for different translations are the same (with only a translation difference between the synthesized feature maps). As a result, the projection only needs to be conducted once. Then, a fast spatial correlation using Pytorch/TensorFlow built-in layer can be used to implement the dense search, enabling high computational efficiency.

In contrast, S2G projected features for different translations are different, and these differences are crucial for translation estimation. Therefore, S2G projection needs to be conducted as many times as the candidate translation poses when implementing dense search, leading to unaffordable computation and memory consumption. This is also why we cannot compare the performance of G2S and S2G synthesis on location estimation using the dense search mechanism.


\begin{table*}[t!]
\setlength{\abovecaptionskip}{0pt}
\setlength{\belowcaptionskip}{0pt}
\setlength{\tabcolsep}{2pt}
\centering
\footnotesize
\caption{\small Performance comparison on KITTI with increasing orientation noise. }
\begin{tabular}{c|c|ccc|ccc|ccc|ccc|ccc|ccc}
\toprule
      \multirow{2}{*}{\begin{tabular}[c]{@{}c@{}}Rotation \\ Noise\end{tabular}}   &   & \multicolumn{3}{c|}{Lateral}                          & \multicolumn{3}{c|}{Longitudinal}                         & \multicolumn{3}{c|}{Azimuth}                      & \multicolumn{3}{c|}{Lateral}                          & \multicolumn{3}{c|}{Longitudinal}                         & \multicolumn{3}{c}{Azimuth}                      \\
       &    & $d=1$          & $d=3$          & $d=5$          & $d=1$         & $d=3$          & $d=5$          & $\theta=1$     & $\theta=3$     & $\theta=5$     & $d=1$          & $d=3$          & $d=5$          & $d=1$         & $d=3$          & $d=5$          & $\theta=1$     & $\theta=3$     & $\theta=5$     \\\midrule
       &   & \multicolumn{9}{c|}{Test1}                                                                                                                             & \multicolumn{9}{c}{Test2}                                                                                                                             \\ \midrule

\multirow{2}{*}{20$^\circ$} & S2G  & 70.02                  & 94.96                  & 97.80                  & 16.22                  & 38.14                  & 48.50                  & 87.57                 & 99.95                 & \textbf{100.00}       & 56.72                  & \textbf{89.63}         & \textbf{93.69}         & 12.76                  & 30.81                  & 41.63                  & 87.52                 & 99.89                 & \textbf{100.00}       \\
                    & G2S (Ours) & \textbf{76.44}         & \textbf{96.34}         & \textbf{98.89}         & \textbf{23.54}         & \textbf{50.57}         & \textbf{62.18}         & \textbf{99.10}        & \textbf{100.00}       & \textbf{100.00}       & \textbf{57.72}         & 86.77                  & 91.16                  & \textbf{14.15}         & \textbf{34.59}         & \textbf{45.00}         & \textbf{98.98}        & \textbf{100.00}       & \textbf{100.00}       \\ \midrule
\multirow{2}{*}{80$^\circ$} & S2G  & 54.17                  & 88.07                  & 94.17                  & 13.46                  & 33.08                  & 43.94                  & 19.56                 & 51.23                 & 70.13                 & 46.90                  & 84.65                  & 92.15                  & 11.92                  & 28.52                  & 38.86                  & 21.82                 & 54.83                 & 74.18                 \\
                    & G2S (Ours)  & \textbf{70.21}         & \textbf{95.47}         & \textbf{98.28}         & \textbf{22.29}         & \textbf{48.90}         & \textbf{59.50}         & \textbf{53.27}        & \textbf{93.98}        & \textbf{98.99}        & \textbf{56.97}         & \textbf{87.72}         & \textbf{92.35}         & \textbf{15.17}         & \textbf{35.39}         & \textbf{47.02}         & \textbf{58.68}        & \textbf{95.92}        & \textbf{99.15}       
\\ \bottomrule
\end{tabular}
\label{tab:G2S_S2G}
\end{table*}


\section{Additional Model Analysis}


Here, we conduct further ablation studies to demonstrate the effectiveness and necessity of each component in the proposed method. Starting from the deep LM~[{\color{green}29}] for joint rotation and translation optimization, we first replace the deep LM optimizer with the proposed Neural pose Optimizer (Neural Opt.). The first two rows of Tab.~\ref{tab:supple_abla} indicate that the proposed neural pose optimizer achieves significantly better performance in rotation estimation but worse performance in translation estimation. 
This demonstrates our intuition that the translation on the neural optimizer input will be absorbed by the deep high-level features inside the optimizer, thus predicting inaccurate translation estimates. 

Next, we construct a decoupling framework where the neural pose optimizer is employed for rotation estimation while the deep LM optimizer is adopted for translation optimization. 
The third row of Tab.~\ref{tab:supple_abla} shows the neural optimizer's powerful rotation estimation performance is mainly inherited in the decoupling framework, benefiting from which the deep LM's lateral translation optimization performance improves.
This demonstrates that it is beneficial to develop different rotation and translation estimation strategies to maximize their performance. 

Then, we replace the deep LM optimizer in the decoupling framework with the proposed Dense Search (DS) mechanism (spatial correlation between synthesized overhead view feature map and observed satellite image feature map). 
From the fourth row of Tab.~\ref{tab:supple_abla}, it can be seen the performance on both rotation and translation increases, indicating the superiority of the proposed dense search mechanism over deep LM optimization. 
The last two rows of Tab.~\ref{tab:supple_abla} present the performance when employing the proposed geometry-guided cross-view transformer for overhead view feature synthesis and including the uncertainty map in the dense search process, respectively. 
Both of them contribute to better performance. 

Furthermore, we conduct experiments to demonstrate whether it is necessary to apply supervision on the estimated translation by the neural optimizer. 
Tab.~\ref{tab:trans_supervision} shows the results. 
It can be seen that the neural optimizer's rotation estimation performance is poor when its translation predictions are not supervised. 
This indicates that additional supervision on the neural optimizer translation predictions provides more clues for the neural optimizer weight learning. 
This is especially important because the feature extractors and the neural optimizer are randomly initialized. Providing more supervision constraints the network learning freedom. 



\begin{table}[ht!]
\setlength{\abovecaptionskip}{0pt}
\setlength{\belowcaptionskip}{0pt}
\setlength{\tabcolsep}{1.5pt}
\centering
\footnotesize
\caption{\small Additional ablation study results of our method on KITTI.}
\begin{tabular}{c|c|c|ccH|ccH|ccH|ccH|ccH|ccH}
\toprule
\multirow{3}{*}{\begin{tabular}[c]{@{}c@{}}Overhead-view\\ Feature\\ Synthesis\end{tabular}}      & \multirow{3}{*}{\begin{tabular}[c]{@{}c@{}}Optimization\\ Scheme\end{tabular}} & \multirow{3}{*}{Uncertainty} & \multicolumn{9}{c|}{Test1}                                                                             & \multicolumn{9}{c}{Test2}                                                                             \\
                                                                                                  &                                                                          &                              & \multicolumn{3}{c|}{Lateral} & \multicolumn{3}{c|}{Longitudinal} & \multicolumn{3}{c|}{Azimuth}          & \multicolumn{3}{c|}{Lateral} & \multicolumn{3}{c|}{Longitudinal} & \multicolumn{3}{c}{Azimuth}          \\ 
                                                                                                  &                                                                          &                              & $d=1$   & $d=3$   & $d=5$   & $d=1$     & $d=3$     & $d=5$    & $\theta=1$ & $\theta=3$ & $\theta=5$ & $d=1$   & $d=3$   & $d=5$   & $d=1$     & $d=3$     & $d=5$    & $\theta=1$ & $\theta=3$ & $\theta=5$ \\ \midrule
\multirow{4}{*}{\begin{tabular}[c]{@{}c@{}}Geometry \\ projection\end{tabular}}                   & LM ($\mathbf{R}$ \& $\mathbf{t}$)                                         & N/A                          & 27.72   & 59.98   & 71.91   & 5.75      & 16.8      & 26.13    & 18.13      & 48.77      & 69.26      & 27.82   & 59.79   & 72.89   & 5.75      & 16.36     & 26.48    & 18.42      & 49.72      & 71         \\
                                                                                                  & Neural Opt. ($\mathbf{R}$ \& $\mathbf{t}$)                                & N/A                          & 4.90    & 15.00   & 25.28   & 4.88      & 15.66     & 25.39    & 97.75      & 100.00     & 100.00     & 5.09    & 15.39   & 25.63   & 5.32      & 15.87     & 25.56    & 97.75      & 100.00     & 100.00     \\
                                                                                                  & Neural Opt. ($\mathbf{R}$) + LM ($\mathbf{t}$)                            & N/A                          & 41.82   & 76.28   & 84.71   & 5.27      & 15.74     & 26.56    & 87.01      & 100.00     & 100.00     & 28.53   & 64.82   & 77.33   & 5.42      & 16.11     & 26.4     & 87.79      & 100.00     & 100.00     \\
                                                                                                  & Neural Opt. ($\mathbf{R}$) + DS ($\mathbf{t}$)                            & No                           & 64.83   & 92.15   & 95.71   & 14.47     & 33.87     & 44.61    & 99.92      & 100.00     & 100.00     & 51.70   & 79.79   & 85.47   & 9.45      & 24.11     & 33.03    & 99.80      & 100.00     & 100.00     \\ \midrule
\multirow{2}{*}{\begin{tabular}[c]{@{}c@{}}Geometry-guided\\ Cross-view Transformer\end{tabular}} & Neural Opt. ($\mathbf{R}$) + DS ($\mathbf{t}$)                            & No                          & 70.29          & 94.30          & 97.35          & 18.29          & 40.39          & 50.01          & 84.44          & 99.84           & \textbf{100.00} & 54.53          & 85.44          & 89.88          & 12.50          & 29.65          & 39.59          & 84.16          & 99.80           & \textbf{100.00} \\
                                                                                                  & Neural Opt. ($\mathbf{R}$) + DS ($\mathbf{t}$)                            & Yes                          & \textbf{76.44} & \textbf{96.34} & \textbf{98.89} & \textbf{23.54} & \textbf{50.57} & \textbf{62.18} & \textbf{99.10} & \textbf{100.00} & \textbf{100.00} & \textbf{57.72} & \textbf{86.77} & \textbf{91.16} & \textbf{14.15} & \textbf{34.59} & \textbf{45.00} & \textbf{98.98} & \textbf{100.00} & \textbf{100.00}
\\\bottomrule
\end{tabular}
\label{tab:supple_abla}
\end{table}


\begin{table}[]
\setlength{\abovecaptionskip}{0pt}
\setlength{\belowcaptionskip}{0pt}
\setlength{\tabcolsep}{1.5pt}
\centering
\footnotesize
\caption{\small Performance of our method with or without translation supervision applied on the proposed neural pose optimizer on KITTI.}
\begin{tabular}{c|ccc|ccc|ccc|ccc|ccc|ccc}
\toprule
\multirow{3}{*}{\begin{tabular}[c]{@{}c@{}}Translation supervision \\ on \\ neural pose optimizer\end{tabular}} & \multicolumn{9}{c|}{Test1}                                                                             & \multicolumn{9}{c}{Test2}                                                                             \\
                                                                                                               & \multicolumn{3}{c|}{Lateral} & \multicolumn{3}{c|}{Longitudinal} & \multicolumn{3}{c|}{Azimuth}          & \multicolumn{3}{c|}{Lateral} & \multicolumn{3}{c|}{Longitudinal} & \multicolumn{3}{c}{Azimuth}          \\
                                                                                                               & $d=1$   & $d=3$   & $d=5$   & $d=1$     & $d=3$     & $d=5$    & $\theta=1$ & $\theta=3$ & $\theta=5$ & $d=1$   & $d=3$   & $d=5$   & $d=1$     & $d=3$     & $d=5$    & $\theta=1$ & $\theta=3$ & $\theta=5$ \\ \midrule
No                                                                                                             & 50.44   & 20.59   & 93.90   & 46.75     & 98.41     & 57.86    & 10.52      & 30.19      & 50.52      & 41.16   & 13.23   & 82.91   & 30.93     & 90.73     & 41.59    & 10.10      & 29.94      & 50.61      \\
Yes                                                                                                            & 76.44   & 96.34   & 98.89   & 23.54     & 50.57     & 62.18    & 99.10      & 100.00     & 100.00     & 56.97   & 87.72   & 92.35   & 15.17     & 35.39     & 47.02    & 58.68      & 95.92      & 99.15     
     \\ \bottomrule
\end{tabular}
\label{tab:trans_supervision}
\end{table}




\section{Number of Iterations of the Proposed Neural Pose Optimizer}

In this section, we study the choice of iteration number for the proposed neural optimizer. Results are presented in Tab.~\ref{tab:iters}. It can be seen that using two iterations contributes to a better performance than using one iteration. However, the performance becomes robust/similar when increasing the iteration number further. Thus, we use the iteration number as two in our proposed framework.

\begin{table*}[ht]
\setlength{\abovecaptionskip}{0pt}
\setlength{\belowcaptionskip}{0pt}
\setlength{\tabcolsep}{2pt}
\centering
\footnotesize
\caption{\small Performance comparison on KITTI with different iteration numbers of the proposed neural optimizer. }
\begin{tabular}{c|ccc|ccc|ccc|ccc|ccc|ccc}
\toprule
\multirow{3}{*}{No. Iterations} & \multicolumn{9}{c|}{Test1}                                                                             & \multicolumn{9}{c}{Test2}                                                                             \\
                           & \multicolumn{3}{c|}{Lateral} & \multicolumn{3}{c|}{Longitudinal} & \multicolumn{3}{c|}{Azimuth}          & \multicolumn{3}{c|}{Lateral} & \multicolumn{3}{c|}{Longitudinal} & \multicolumn{3}{c}{Azimuth}          \\
                           & $d=1$   & $d=3$   & $d=5$   & $d=1$     & $d=3$     & $d=5$    & $\theta=1$ & $\theta=3$ & $\theta=5$ & $d=1$   & $d=3$   & $d=5$   & $d=1$     & $d=3$     & $d=5$    & $\theta=1$ & $\theta=3$ & $\theta=5$ \\ \midrule
1                          & 75.19   & 95.68   & 97.93   & 21.23     & 47.15     & 57.96    & 99.87      & 100.00     & 100.00     & 53.94   & 84.23   & 89.72   & 13.52     & 33.90     & 45.35    & 99.80      & 100.00     & 100.00     \\
2                          & 76.44   & 96.34   & 98.89   & 23.54     & 50.57     & 62.18    & 99.10      & 100.00     & 100.00     & 57.72   & 86.77   & 91.16   & 14.15     & 34.59     & 45.00    & 98.98      & 100.00     & 100.00     \\
3                          & 75.85   & 96.16   & 98.54   & 23.32     & 50.68     & 61.17    & 80.52      & 99.89      & 100.00     & 58.04   & 86.58   & 90.77   & 13.80     & 33.43     & 44.78    & 77.53      & 100.00     & 100.00     \\
4                          & 76.60   & 96.50   & 98.75   & 23.32     & 50.41     & 60.69    & 79.91      & 99.95      & 100.00     & 58.21   & 87.88   & 92.12   & 15.02     & 35.93     & 46.91    & 77.34      & 100.00     & 100.00     \\
5                          & 75.30   & 95.73   & 98.33   & 22.71     & 50.78     & 61.17    & 81.13      & 99.97      & 100.00     & 58.98   & 88.50   & 92.54   & 15.02     & 35.85     & 46.80    & 78.23      & 99.88      & 100.00     \\\bottomrule
\end{tabular}
\label{tab:iters}
\end{table*}

\section{Different Initial Values}

\paragraph{Orientation.}
Below, we increase the rotation noise and compare the performance of our method with deep LM. 
The location search range follows the official setting~[{\color{green}29}]: within a 40m $\times$ 40m search space. 
The rotation noise is set to $20^\circ, 40^\circ$, and $80^\circ$. 
From the results in Tab.~\ref{tab:kitti_increase_Rnoise}, 
it can be seen that deep LM almost fails on rotation estimation when the rotation noise increases to $80^\circ$, with only $9\%$ of the images whose estimated rotation is within $3^\circ$ to its ground truth values. 
In contrast, our method makes over $90\%$ of the estimated rotation within $3^\circ$ of their GT values. 
Since the rotation ambiguity has been significantly reduced, the translation estimation performance of our method is robust. 

\begin{figure}
    \centering
    \begin{minipage}{0.75\linewidth}
        \includegraphics[width=0.48\linewidth, height=0.24\linewidth]{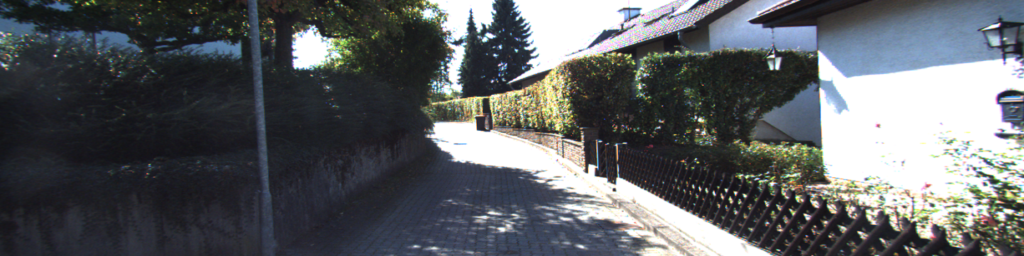}
        \adjincludegraphics[width=0.24\linewidth,trim={{0.23\width} {0.23\width} {0.23\width} {0.23\width}},clip]{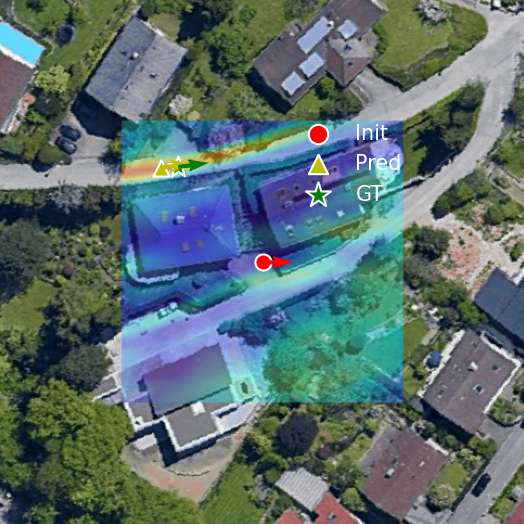}
        \adjincludegraphics[width=0.24\linewidth,trim={{0.23\width} {0.23\width} {0.23\width} {0.23\width}},clip]{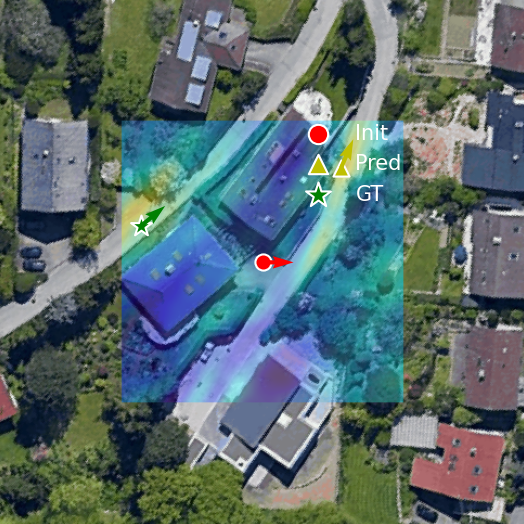} \\
        \includegraphics[width=0.48\linewidth, height=0.24\linewidth]{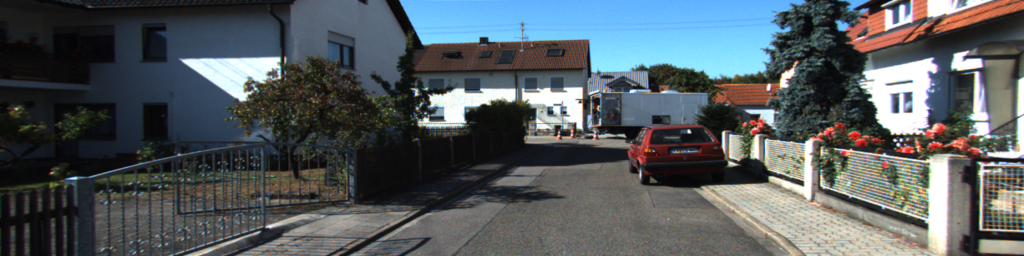}
        \adjincludegraphics[width=0.24\linewidth,trim={{0.23\width} {0.23\width} {0.23\width} {0.23\width}},clip]{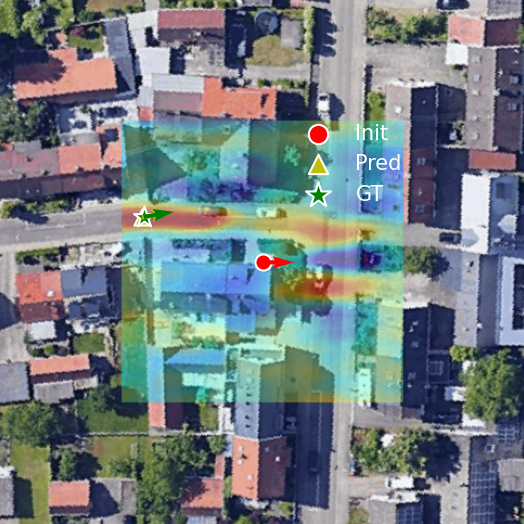}
        \adjincludegraphics[width=0.24\linewidth,trim={{0.23\width} {0.23\width} {0.23\width} {0.23\width}},clip]{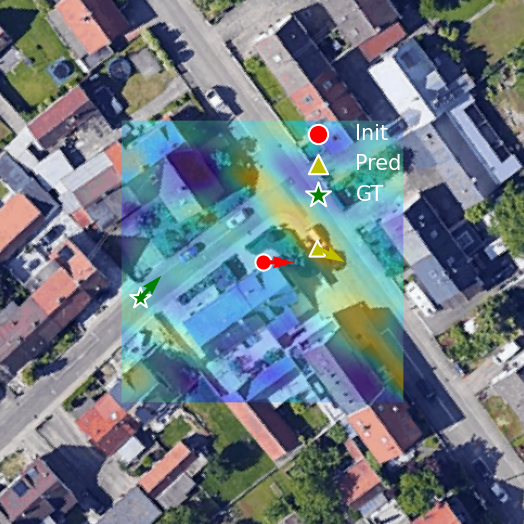} \\
        \begin{minipage}{0.48\linewidth}
            \centerline{\small Query Image}
        \end{minipage}
        \begin{minipage}{0.24\linewidth}
            \centerline{\small $20^\circ$ Rotation Ambiguity}
        \end{minipage}
        \begin{minipage}{0.24\linewidth}
            \centerline{\small $100^\circ$ Rotation Ambiguity}
        \end{minipage}\\
    \end{minipage}
    \caption{Localization results of our method when rotation ambiguity is different. The satellite images in each row are from the same place with different rotations to mimic the different rotation noises of the ground camera. When the orientation noise is large ($100^\circ$), the rotation estimation ambiguity becomes significant because the limited scene content captured by a query image can be matched to the different parts on a satellite image.}
    \label{fig:rot_noise_loc_visualization}
\end{figure}

Tab.~\ref{tab:all_rot_noise} provides the quantitative evaluation of our method when increasing the rotation noise to $180^\circ$. 
It can be seen that the rotation estimation performance drops when the rotation noise increases to a large number, which also affects the translation estimation performance.
Fig.~\ref{fig:rot_noise_loc_visualization} shows two examples. 
The query images in Fig.~\ref{fig:rot_noise_loc_visualization} tell that the cameras are facing towards the road. 
When the rotation prior is relatively accurate, we can quickly determine the camera's orientation by comparing the rotation angles between the initialized orientation and the road directions. Then, the camera location will also be estimated on the right road, as shown in the middle column of Fig.~\ref{fig:rot_noise_loc_visualization}. 
However, when the rotation ambiguity is considerable, there is a large probability of the query image being matched to a wrong part of the satellite image, resulting incorrect rotation and translation estimates, as shown in the right column of Fig.~\ref{fig:rot_noise_loc_visualization}. 
In practice, we suggest first restricting the rotation ambiguity of the query cameras to an acceptable range(\eg, within $80^\circ$) and then adopting ground-to-satellite image matching to refine this pose.  


\begin{table*}[ht]
\setlength{\abovecaptionskip}{0pt}
\setlength{\belowcaptionskip}{0pt}
\setlength{\tabcolsep}{2pt}
\centering
\footnotesize
\caption{\small Performance comparison on KITTI with increasing orientation noise. }
\begin{tabular}{c|c|ccc|ccc|ccc|ccc|ccc|ccc}
\toprule
      \multirow{2}{*}{\begin{tabular}[c]{@{}c@{}}Rotation \\ Noise\end{tabular}}   &   & \multicolumn{3}{c|}{Lateral}                          & \multicolumn{3}{c|}{Longitudinal}                         & \multicolumn{3}{c|}{Azimuth}                      & \multicolumn{3}{c|}{Lateral}                          & \multicolumn{3}{c|}{Longitudinal}                         & \multicolumn{3}{c}{Azimuth}                      \\
       &    & $d=1$          & $d=3$          & $d=5$          & $d=1$         & $d=3$          & $d=5$          & $\theta=1$     & $\theta=3$     & $\theta=5$     & $d=1$          & $d=3$          & $d=5$          & $d=1$         & $d=3$          & $d=5$          & $\theta=1$     & $\theta=3$     & $\theta=5$     \\\midrule
       &   & \multicolumn{9}{c|}{Test1}                                                                                                                             & \multicolumn{9}{c}{Test2}                                                                                                                             \\ \midrule

\multirow{2}{*}{20$^\circ$} & LM~[{\color{green}29}]   & 35.54          & 70.77          & 80.36          & 5.22           & 15.88          & 26.13          & 19.64          & 51.76           & 71.72           & 27.82          & 59.79          & 72.89          & 5.75           & 16.36          & 26.48          & 18.42          & 49.72           & 71.00           \\
                    & Ours & \textbf{76.44} & \textbf{96.34} & \textbf{98.89} & \textbf{23.54} & \textbf{50.57} & \textbf{62.18} & \textbf{99.10} & \textbf{100.00} & \textbf{100.00} & \textbf{57.72} & \textbf{86.77} & \textbf{91.16} & \textbf{14.15} & \textbf{34.59} & \textbf{45.00} & \textbf{98.98} & \textbf{100.00} & \textbf{100.00} \\ \midrule
\multirow{2}{*}{40$^\circ$} & LM~[{\color{green}29}]   & 32.02          & 68.43          & 79.25          & 5.46           & 16.57          & 27.46          & 13.70          & 36.47           & 52.74           & 26.58          & 62.94          & 74.93          & 5.34           & 16.12          & 26.47          & 10.82          & 32.38           & 48.69           \\
                    & Ours & \textbf{70.42} & \textbf{94.75} & \textbf{97.43} & \textbf{20.49} & \textbf{48.50} & \textbf{58.49} & \textbf{69.12} & \textbf{99.68}  & \textbf{99.97}  & \textbf{55.52} & \textbf{86.20} & \textbf{91.10} & \textbf{13.66} & \textbf{33.74} & \textbf{44.95} & \textbf{76.45} & \textbf{99.59}  & \textbf{99.99}  \\ \midrule
\multirow{2}{*}{80$^\circ$} & LM~[{\color{green}29}]  &26.95	&62.39	&78.40	&5.14	&15.69	&26.27	&3.10	&8.88	&15.00  & 22.43	&54.63	&71.03	&5.17	&15.78	&25.97	&3.05	&8.50	&14.25            \\
                    & Ours & \textbf{70.21} & \textbf{95.47} & \textbf{98.28} & \textbf{22.29} & \textbf{48.90} & \textbf{59.50} & \textbf{53.27} & \textbf{93.98}  & \textbf{98.99}  & \textbf{56.97} & \textbf{87.72} & \textbf{92.35} & \textbf{15.17} & \textbf{35.39} & \textbf{47.02} & \textbf{58.68} & \textbf{95.92}  & \textbf{99.15}  
\\ \bottomrule
\end{tabular}
\label{tab:kitti_increase_Rnoise}
\end{table*}

\begin{table*}[ht]
\setlength{\abovecaptionskip}{0pt}
\setlength{\belowcaptionskip}{0pt}
\setlength{\tabcolsep}{2.5pt}
\centering
\footnotesize
\caption{\small Performance of our method on KITTI with different rotation noises.}
\begin{tabular}{c|ccc|ccc|ccc|ccc|ccc|ccc}
\toprule
\multirow{3}{*}{\begin{tabular}[c]{@{}c@{}}Rotation\\ Ambiguity\end{tabular}} & \multicolumn{9}{c|}{Test1}                                                                             & \multicolumn{9}{c}{Test2}                                                                                                                                                                                                                                                \\
                                                                             & \multicolumn{3}{c|}{Lateral} & \multicolumn{3}{c|}{Longitudinal} & \multicolumn{3}{c|}{Azimuth}          & \multicolumn{3}{c|}{Lateral}                                                       & \multicolumn{3}{c|}{Longitudinal}                                                  & \multicolumn{3}{c}{Azimuth}                                                                      \\
                                                                             & $d=1$   & $d=3$   & $d=5$   & $d=1$     & $d=3$     & $d=5$    & $\theta=1$ & $\theta=3$ & $\theta=5$ & {$d=1$} & {$d=3$} & {$d=5$} & {$d=1$} & {$d=3$} & {$d=5$} & {$\theta=1$} & {$\theta=3$} & {$\theta=5$} \\ \midrule
20$^\circ$                                                                           & 76.44   & 96.34   & 98.89   & 23.54     & 50.57     & 62.18    & 99.10      & 100.00     & 100.00     & 57.72                     & 86.77                     & 91.16                     & 14.15                     & 34.59                     & 45.00                     & 98.98                          & 100.00                         & 100.00                         \\
40$^\circ$                                                                            & 70.42   & 94.75   & 97.43   & 20.49     & 48.50     & 58.49    & 69.12      & 99.68      & 99.97      & 55.52                     & 86.20                     & 91.10                     & 13.66                     & 33.74                     & 44.95                     & 76.45                          & 99.59                          & 99.99                          \\
60$^\circ$                                                                            & 72.78   & 95.02   & 98.07   & 21.60     & 47.79     & 58.20    & 60.75      & 97.80      & 99.84      & 54.85                     & 85.76                     & 90.86                     & 14.12                     & 33.36                     & 45.12                     & 67.78                          & 98.36                          & 99.76                          \\
80$^\circ$                                                                            & 70.21   & 95.47   & 98.28   & 22.29     & 48.90     & 59.50    & 53.27      & 93.98      & 98.99      & 56.97                     & 87.72                     & 92.35                     & 15.17                     & 35.39                     & 47.02                     & 58.68                          & 95.92                          & 99.15                          \\
100$^\circ$                                                                           & 50.78   & 83.28   & 91.44   & 16.94     & 39.04     & 49.83    & 20.67      & 51.55      & 67.00      & 40.82                     & 75.10                     & 85.32                     & 13.59                     & 32.37                     & 42.91                     & 18.97                          & 50.98                          & 68.39                          \\
120$^\circ$                                                                           & 37.26   & 71.38   & 85.13   & 14.92     & 34.96     & 45.51    & 13.07      & 34.53      & 46.57      & 30.69                     & 63.96                     & 77.49                     & 11.31                     & 28.52                     & 38.44                     & 11.48                          & 30.67                          & 43.46                          \\
140$^\circ$                                                                           & 29.42   & 62.10   & 77.52   & 10.39     & 25.87     & 35.70    & 7.55       & 20.70      & 31.25      & 22.83                     & 52.74                     & 69.24                     & 8.51                      & 21.04                     & 29.75                     & 6.93                           & 18.56                          & 26.72                          \\
160$^\circ$                                                                          & 24.83   & 56.19   & 73.63   & 10.52     & 26.45     & 36.05    & 5.78       & 17.52      & 25.68      & 19.08                     & 46.43                     & 62.19                     & 8.15                      & 20.70                     & 30.22                     & 4.53                           & 13.46                          & 21.81                          \\
180$^\circ$                                                                           & 16.94   & 45.77   & 63.80   & 7.77      & 20.81     & 29.92    & 3.47       & 9.70       & 16.09      & 14.86                     & 39.54                     & 57.44                     & 7.11                      & 18.96                     & 27.46                     & 2.56                           & 8.17                           & 14.00   \\ \bottomrule                      
\end{tabular}
\label{tab:all_rot_noise}
\end{table*}

\paragraph{Location.}
Next, we investigate the performance of our method with different location initialization ranges.  Tab.~\ref{tab:loc_priors_comparison} presents the comparison between our approach and deep LM~[{\color{green}29}]. 
It can be seen that both methods achieve better performance when the location initialization range is smaller.
The performance gap between deep LM and our method increases as the location initialization range increases. 
This indicates our method is more robust to the location initialization ranges/errors than LM because of the proposed dense search strategy for location estimation.   
Tab.~\ref{tab:loc_priors_ours} provides the results of our method when we keep increasing the location search range. 
The performance decreases gradually as the location search range increases. 
But even with a search range of 100m $\times$ 100m, our method still outperforms LM when its location search range is 20m $\times$ 20m.


\begin{table}[]
\setlength{\abovecaptionskip}{0pt}
\setlength{\belowcaptionskip}{0pt}
\setlength{\tabcolsep}{1.5pt}
\centering
\footnotesize
\caption{\small Performance comparison between LM[{\color{green}29}] and our method with different location priors and $20^\circ$ orientation noise. }
\begin{tabular}{c|c|ccc|ccc|ccc|ccc|ccc|ccc}
\toprule
\multirow{3}{*}{\begin{tabular}[c]{@{}c@{}}Initialization\\ Range\end{tabular}}                        & \multirow{3}{*}{Methods} & \multicolumn{9}{c|}{Test1}                                                                             & \multicolumn{9}{c}{Test2}                                                                             \\
                                                                                                &                          & \multicolumn{3}{c|}{Lateral} & \multicolumn{3}{c|}{Longitudinal} & \multicolumn{3}{c|}{Azimuth}          & \multicolumn{3}{c|}{Lateral} & \multicolumn{3}{c|}{Longitudinal} & \multicolumn{3}{c}{Azimuth}          \\
                                                                                                &                          & $d=1$   & $d=3$   & $d=5$   & $d=1$     & $d=3$     & $d=5$    & $\theta=1$ & $\theta=3$ & $\theta=5$ & $d=1$   & $d=3$   & $d=5$   & $d=1$     & $d=3$     & $d=5$    & $\theta=1$ & $\theta=3$ & $\theta=5$ \\ \midrule
\multirow{2}{*}{10m x 10m}                                                                      & LM [{\color{green}29}]       & 64.86   & 92.23   & 96.98   & 29.08     & 69.49     & 88.66    & 36.92      & 73.95      & 86.88      & 55.98   & 90.84   & 96.43   & 25.97     & 66.96     & 88.12    & 31.36      & 69.46      & 84.50      \\ 
                                                                                                & Ours                     & \textbf{89.37}   & \textbf{98.67}   & \textbf{99.81}   & \textbf{26.72}     & \textbf{56.03}     & \textbf{72.36}    & \textbf{98.20}      & \textbf{100.00}     & \textbf{100.00}     & \textbf{62.16}   & \textbf{20.49}   & \textbf{89.87}   & \textbf{47.97}     & \textbf{96.66}     & \textbf{66.12}    & \textbf{97.44}      & \textbf{99.95}      & \textbf{100.00}     \\ \midrule
\multirow{2}{*}{20m x 20m}                                                                      & LM [{\color{green}29}]    & 44.66   & 73.92   & 81.18   & 12.06     & 35.62     & 54.73    & 25.31      & 57.41      & 74.48      & 34.17   & 72.30   & 81.15   & 11.56     & 35.08     & 53.77    & 11.40      & 48.18      & 65.80      \\
                                                                                                & Ours                     & \textbf{85.85}   & \textbf{98.46}   & \textbf{99.55}   & \textbf{23.27}     & \textbf{46.99}     & \textbf{58.39}    & \textbf{98.89}      & \textbf{99.97}      & \textbf{100.00}     & \textbf{60.01}   & \textbf{14.69}   & \textbf{87.96}   & \textbf{35.64}     & \textbf{92.97}     & \textbf{48.46}    & \textbf{99.42}      & \textbf{100.00}     & \textbf{100.00}     \\\midrule
\multirow{2}{*}{\begin{tabular}[c]{@{}c@{}}40m x 40m\end{tabular}} & LM  [{\color{green}29}]   & 35.54   & 70.77   & 80.36   & 5.22      & 15.88     & 26.13    & 19.64      & 51.76      & 71.72      & 27.82   & 59.79   & 72.89   & 5.75      & 16.36     & 26.48    & 18.42      & 49.72      & 71.00      \\
                                                                                                & Ours                    & \textbf{76.44}   & \textbf{96.34}   & \textbf{98.89}   & \textbf{23.54}     & \textbf{50.57}     & \textbf{62.18}    & \textbf{99.10}      & \textbf{100.00}     & \textbf{100.00}     & \textbf{56.97}   & \textbf{87.72}   & \textbf{92.35}   & \textbf{15.17}     & \textbf{35.39}     & \textbf{47.02}    & \textbf{58.68 }     & \textbf{95.92}      & \textbf{99.15}      \\ \bottomrule
\end{tabular}
\label{tab:loc_priors_comparison}
\end{table}

\begin{table}[]
\setlength{\abovecaptionskip}{0pt}
\setlength{\belowcaptionskip}{0pt}
\setlength{\tabcolsep}{2pt}
\centering
\footnotesize
\caption{\small Performance of our method with different location priors and $20^\circ$ orientation noise.  }
\begin{tabular}{c|ccc|ccc|ccc|ccc|ccc|ccc}
\toprule
\multirow{3}{*}{\begin{tabular}[c]{@{}c@{}}Search\\ Region\end{tabular}} & \multicolumn{9}{c|}{Test1}                                                                             & \multicolumn{9}{c}{Test2}                                                                             \\
                                                                         & \multicolumn{3}{c|}{Lateral} & \multicolumn{3}{c|}{Longitudinal} & \multicolumn{3}{c|}{Azimuth}          & \multicolumn{3}{c|}{Lateral} & \multicolumn{3}{c|}{Longitudinal} & \multicolumn{3}{c}{Azimuth}          \\
                                                                         & $d=1$   & $d=3$   & $d=5$   & $d=1$     & $d=3$     & $d=5$    & $\theta=1$ & $\theta=3$ & $\theta=5$ & $d=1$   & $d=3$   & $d=5$   & $d=1$     & $d=3$     & $d=5$    & $\theta=1$ & $\theta=3$ & $\theta=5$ \\ \midrule
10m x 10m                                                                 & 89.37   & 98.67   & 99.81   & 26.72     & 56.03     & 72.36    & 98.20      & 100.00     & 100.00     & 62.16   & 89.87   & 96.66   & 20.49     & 47.97     & 66.12    & 97.44      & 99.95      & 100.00     \\
20m x 20m                                                                  & 85.85   & 98.46   & 99.55   & 23.27     & 46.99     & 58.39    & 98.89      & 99.97      & 100.00     & 60.01   & 87.96   & 92.97   & 14.69     & 35.64     & 48.46    & 99.42      & 100.00     & 100.00     \\
40m x 40m                                                                  & 76.44   & 96.34   & 98.89   & 23.54     & 50.57     & 62.18    & 99.10      & 100.00     & 100.00     & 57.72   & 86.77   & 91.16   & 14.15     & 34.59     & 45.00    & 98.98      & 100.00     & 100.00     \\
60m x 60m                                                                  & 68.30   & 93.32   & 96.24   & 14.10     & 30.69     & 38.06    & 98.04      & 100.00     & 100.00     & 47.40   & 79.28   & 85.96   & 8.31      & 19.48     & 25.75    & 98.59      & 100.00     & 100.00     \\
80m x 80m                                                                  & 64.22   & 89.40   & 93.53   & 12.06     & 27.46     & 34.46    & 99.73      & 100.00     & 100.00     & 43.98   & 73.57   & 81.49   & 6.58      & 15.62     & 21.44    & 99.87      & 100.00     & 100.00     \\
100m x 100m                                                                & 59.32   & 87.44   & 91.60   & 11.03     & 23.96     & 31.20    & 95.73      & 100.00     & 100.00     & 41.36   & 71.39   & 79.09   & 5.83      & 14.93     & 20.98    & 91.01      & 100.00     & 100.00    \\ \bottomrule
\end{tabular}
\label{tab:loc_priors_ours}
\end{table}

\section{Weakness Discussions}



\begin{figure*}[t]
\setlength{\abovecaptionskip}{0pt}
\setlength{\belowcaptionskip}{0pt}
    \centering
    \begin{minipage}{\linewidth}
    \centering
        \begin{minipage}{0.3\linewidth}
        \includegraphics[width=\linewidth, height=0.5\linewidth]{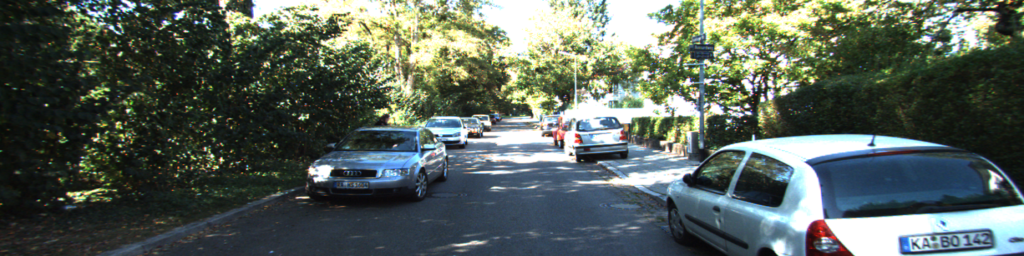}
        \centerline{\small Query Image}
    \end{minipage}
    \begin{minipage}{0.15\linewidth}
        \adjincludegraphics[width=\linewidth,trim={{0.23\width} {0.23\width} {0.23\width} {0.23\width}},clip]{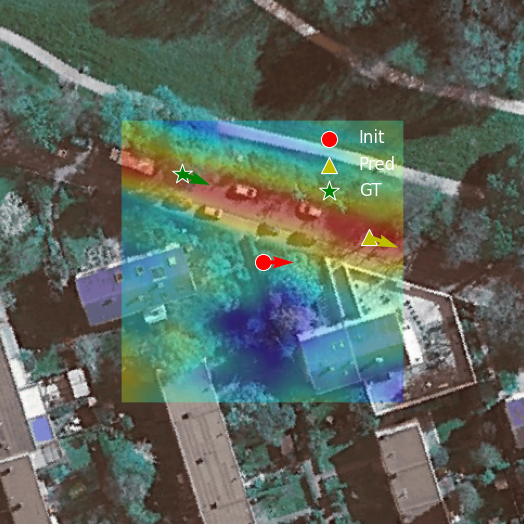}
        \centerline{\small Prediction}
    \end{minipage}
    \hspace{0.5em}
    \begin{minipage}{0.3\linewidth}
        \includegraphics[width=\linewidth, height=0.5\linewidth]{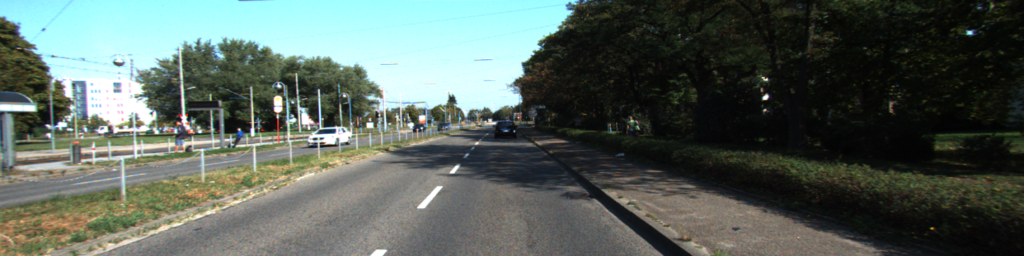}
        \centerline{\small Query Image}
    \end{minipage}
    \begin{minipage}{0.15\linewidth}
        \adjincludegraphics[width=\linewidth,trim={{0.23\width} {0.23\width} {0.23\width} {0.23\width}},clip]{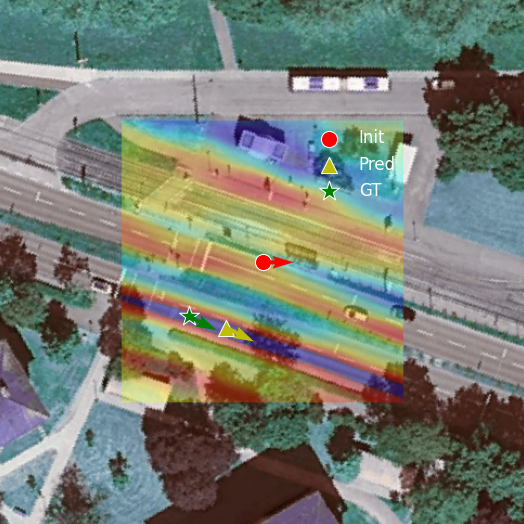}
        \centerline{\small Prediction}
    \end{minipage}
    \end{minipage}
    \caption{Longitudinal pose estimation ambiguity by ground-to-satellite image matching. Determining the camera's position along the road (longitudinal pose) is hard because the scenes along the driving direction are rather monotonous, but the lateral pose can be easily estimated. }. 
    \label{fig:failure}
\end{figure*}

As discussed above, as the rotation and translation ambiguity increases, the localization performance decreases because the scene content captured by the query image may be similar to the scene contents on different parts of the satellite image. Additionally, there is an inherent ambiguity in longitudinal pose estimation when using a single query image for ground-to-satellite image matching, as shown in Fig.~\ref{fig:failure}. Leveraging a multi-camera system with a $360^\circ$ field of view or a continuous video could potentially improve the informativeness of the query place and thus improve the localization performance. 


\end{document}